%% file: 0_main.tex
\documentclass[sigconf]{acmart}
\usepackage{extarrows}
\usepackage{color} 
\usepackage{xcolor}
\usepackage{makecell}
\usepackage{amsmath,amsfonts}
\usepackage{algorithmic}
\usepackage{array}
\usepackage{graphicx}
\usepackage{stfloats}
\usepackage{multicol}
\usepackage{multirow}
\usepackage{hhline}
\usepackage{colortbl}
\usepackage{bm}
\usepackage{url}
\usepackage{enumitem}
\usepackage{CJKutf8}
\usepackage{tabularray}
\usepackage{bigstrut}
\usepackage{booktabs}
\usepackage{arydshln}
\usepackage{utfsym}
\usepackage{hyperref}
\usepackage{float}
\usepackage{threeparttable}
\usepackage{printlen}
\usepackage{algorithm}
\usepackage{algorithmic}
\usepackage{graphicx}
\usepackage{subcaption}
\usepackage{bbding}
\definecolor{lightgray}{rgb}{0.88, 0.92, 0.98}
\definecolor{defblue}{rgb}{0.1843, 0.3333, 0.6}
\definecolor{defred}{rgb}{0.88, 0.2510, 0.3294}

\definecolor{defgreen1}{rgb}{ 0.910,  0.953,  0.855}
\definecolor{defgreen2}{rgb}{0.82,  0.902,  0.710}
\definecolor{defgreen3}{rgb}{0.713,  0.903,  0.648}
\definecolor{defgreen4}{rgb}{ 0.725,  0.855,  0.561}

\definecolor{defyellow}{rgb}{1,  0.983,  0.717}
\definecolor{defyellowtext}{rgb}{1,  0.851,  0.438}
\definecolor{defred3}{rgb}{ 1,  0.398,  0.399}

\definecolor{customgreen}{rgb}{0.3647, 0.6784, 0.3294}

\AtBeginDocument{%
  \providecommand\BibTeX{{%
    \normalfont B\kern-0.5em{\scshape i\kern-0.25em b}\kern-0.8em\TeX}}}

\makeatletter
\renewcommand*{\@fnsymbol}[1]{%
  \ensuremath{%
    \ifcase#1\or 
      \dagger \or              
      \text{\Envelope} \or     
      \ddagger \or \S \or \P \else\@ctrerr
    \fi
  }%
}
\makeatother

\acmConference[MM '26]{Proceedings of the 34th ACM International Conference on Multimedia}{November 10--14, 2026}{Rio de Janeiro, Brazil}
\acmBooktitle{Proceedings of the 34th ACM International Conference on Multimedia (MM '26), November 10--14, 2026, Rio de Janeiro, Brazil}
\acmDOI{10.1145/3767308.3835461}
\acmISBN{979-8-4007-2213-4/2026/11}
\renewcommand\footnotetextcopyrightpermission[1]{}
\setcopyright{none}

\begin{document}
\begin{CJK}{UTF8}{gbsn}
\begin{sloppypar}
\title{LightAIR: Lightweight Action Inversion and Riemannian Rectification for Text-based Person Anomaly Search}

\author{Yulun Zhang}
\orcid{0009-0003-1119-7232}
\affiliation{
  \institution{\normalsize Shenzhen Institutes of Advanced Technology, Chinese Academy of Sciences}
  \city{Shenzhen}
  \country{China}
  }
\email{rainylondon.z@gmail.com}

\author{Zixu Li}
\orcid{0009-0001-5136-159X}
\authornote{Project Leader: Zixu Li.} 
\affiliation{
  \institution{\normalsize Shandong University}
  \city{Jinan}
  \country{China}
}
\email{lizixu.cs@gmail.com}

\author{Zhiwei Chen}
\orcid{0009-0003-0365-8553}
\affiliation{
  \institution{\normalsize Shandong University}
  \city{Jinan}
  \country{China}
  }
\email{zivczw@gmail.com}

\author{Zhiheng Fu}
\orcid{0009-0007-7724-5662}
\affiliation{
  \institution{\normalsize Shandong University}
  \city{Jinan}
  \country{China}
  }
\email{fuzhiheng8@gmail.com}

\author{Wenbo Wang}
\orcid{0009-0009-1440-1685}
\affiliation{
  \institution{\normalsize Shandong University}
  \city{Jinan}
  \country{China}
  }
\email{wangwenbo@mail.sdu.edu.cn}

\author{Zihang Qiu}
\orcid{0009-0009-5335-0868}
\affiliation{
  \institution{\normalsize Shandong University}
  \city{Jinan}
  \country{China}
  }
\email{qiuzihang@mail.sdu.edu.cn}

\author{Zhilin Wang}
\orcid{0009-0001-7101-1220}
\affiliation{
  \institution{\normalsize Shandong University}
  \city{Jinan}
  \country{China}
  }
\email{wangzhilin@mail.sdu.edu.cn}

\author{Ruxin Wang}
\orcid{0000-0003-4772-3284}
\authornote{Corresponding author: Ruxin Wang.} 
\affiliation{
  \institution{\normalsize Shenzhen Institutes of Advanced Technology, Chinese Academy of Sciences}
  \city{Shenzhen}
  \country{China}
}
\email{rx.wang@siat.ac.cn}

\author{Yupeng Hu}
\orcid{0000-0002-5653-8286}
\affiliation{%
  \institution{\normalsize Shandong University}
  \city{Jinan}
  \country{China}
  }
\email{huyupeng@sdu.edu.cn}

\begin{abstract}
Traditional Text-based Person Search (TPS) is typically limited to matching static appearance attributes, severely neglecting dynamic action information. The Text-based Person Anomaly Search (TPAS) task bridges this gap, requiring models to locate micro-level specific abnormal behaviors while matching macro-level appearance of pedestrians.
However, current TPAS methods face fundamental limitations: external explicit pose estimators are fragile in unconstrained surveillance scenarios, and implicit learning encounters visual decoupling failure under pixel-level entanglement, causing dominant appearance information to easily swallow and contaminate subtle action features.
Furthermore, performing contrastive optimization on hard negative samples (``same appearance, different actions'') in conventional Euclidean spaces induces severe shortcut learning.
To address these, we propose the \textbf{Light}weight \textbf{A}ction \textbf{I}nversion and \textbf{R}iemannian rectification network (\textbf{LightAIR}). First, it introduces textual semantic priors as anchors via a lightweight action inversion operator to extract pure action features, thereby overcoming visual-inherent coupling. Subsequently, it employs orthogonal null-space projection to constrain appearance features within the orthogonal complement space of action features, guaranteeing strict forward decoupling. Finally, we designed a gradient rectification module that computes the Riemannian gradient to constrain the backpropagation trajectory, forcing the gradient flow to update strictly along the tangent space that preserves decoupling properties, thereby cutting off harmful shortcuts. 
Extensive experiments on the widely used TPAS and TIPR datasets demonstrate that LightAIR significantly outperforms existing state-of-the-art methods. Codes are available at \textcolor{orange}{\href{https://github.com/rainy-london/LightAIR}{https://github.com/rainy-london/LightAIR}}.

\end{abstract}


\maketitle

\input{1_int.tex}

\input{2_rel.tex}
\input{3_met.tex}

\input{4_exp.tex}
\input{5_con.tex}

\begin{acks}
This paper was supported in part by the National Key R\&D Program of China under Grant 2022YFA1008300, in part by the National Natural Science Foundation of China under Grants 12471308, 62276155, and 62576195, in part by the Key R\&D Program of Shandong Province (Major scientific and technological innovation projects), China, No.: 2025CXGC020101
\end{acks}

\clearpage

\bibliographystyle{ACM-Reference-Format}
\bibliography{reference}
\input{X_suppl}

\balance

\end{sloppypar}
\end{CJK}
\end{document}

%% file: 1_int.tex
\begin{figure}[t]
	\includegraphics[width=1\linewidth]{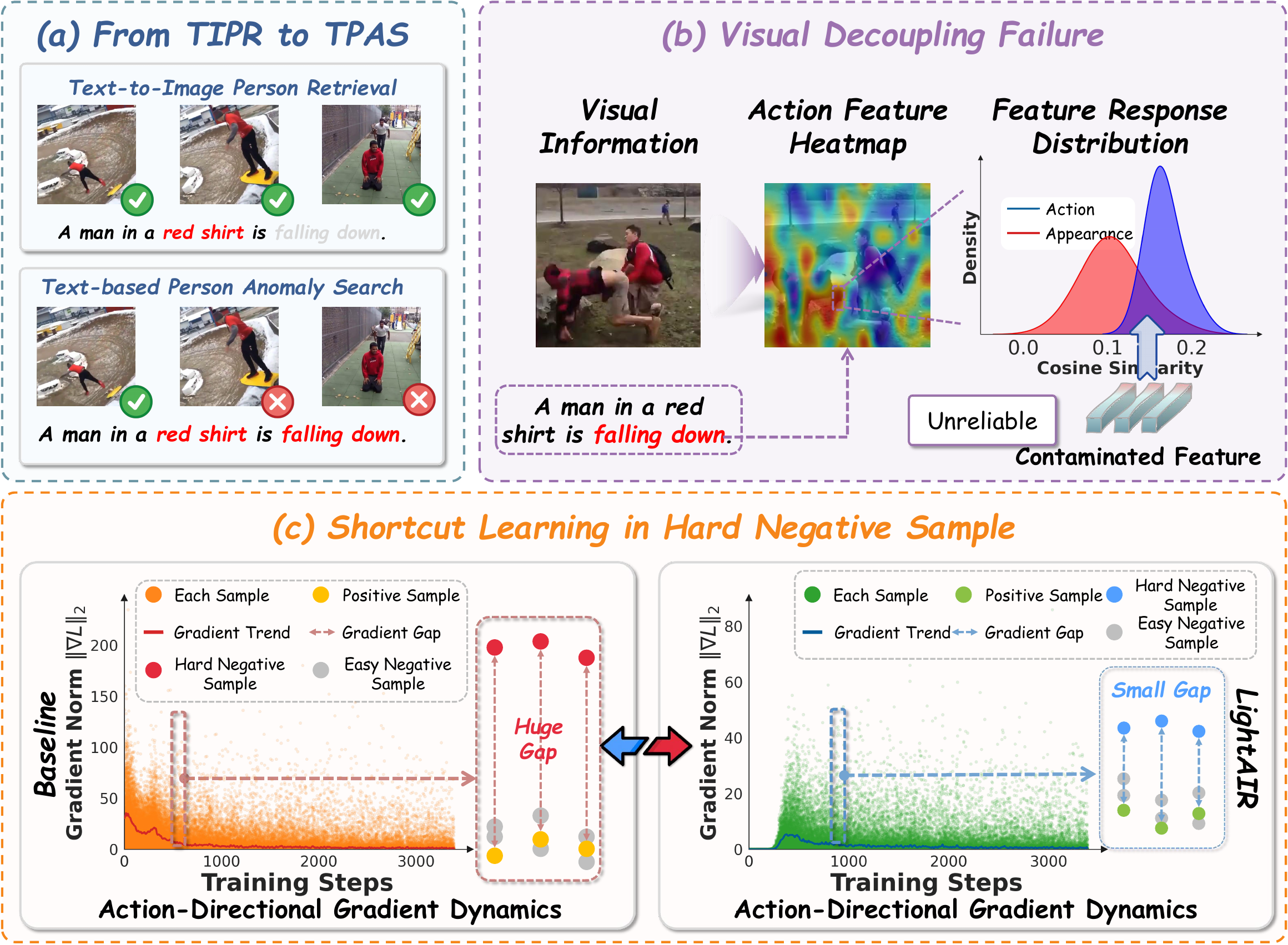}
	\caption{(a) Examples of TIPR and TPAS tasks. (b) Pixel-level entanglement of appearance~\&~action features leads to unreliable feature extraction. (c) Gradient dynamics reveal that massive gradient thrusts induce shortcut learning, whereas LightAIR effectively suppresses this.}
	\label{fig:intro}
\end{figure}

\section{Introduction}

Text-based Person Search (TPS)~\cite{Bi-IRRA, Ufinebench, IRRA} locates specific targets via natural language across multiple cameras in large-scale image databases, offering significant practical value in semantic understanding~\cite{wu2026language,li2026ragtrack,RankVR,R3,HINT,bi2026echorl,MELT,xiao2026layer,xiao2026not,zhao2026hieramp,li2026cadtrack,wu2025sasw}, multimodal retrieval~\cite{sun2023hierarchical,STABLE,qin2023cross,IMAGINE,bi2026the,yuan2025prototype,ReTrack,REFINE,hu2021coarse,hu2023semantic,hu2021video,chen2026task}, and multimodal learning~\cite{11595022,xiao2026promptbased,li2023cross,ERASE,bi2025llava,OmniEgo-R,li2024incomplete,xiao2025visual,yu2025knowledge,huang2026detecting,10720843,zhong2025semi}.
Despite recent TPS progress in vision-language pre-training~\cite{lyu2026hallu_pade,encoder,Bi2025CoTKineticsAT,wang2026ascd,Bi2025PRISMSI,FineCIR,xiao2026reversible,PAIR,MEDIAN,sun2024robust,OFFSET,lyu2025hallu_vdc,lu2026riemannian} and fine-grained attribute learning~\cite{TBPS-CLIP}, traditional methods limit their scope to ``identifying who'' (i.e., static appearance like clothing), severely neglecting dynamic actions detailing ``what they are doing''~\cite{EgoAction,EgoAdapt,lin2026beyond,xiao2026staying,zhao2026seeing}.
In real-world security scenarios, identifying abnormal actions (e.g., falling or being struck) is often more urgent than simple identity verification. Thus, Yang et al.~\cite{CMP2025} proposed Text-based Person Anomaly Search (TPAS), requiring models to match macro-level appearances while precisely locating specific micro-level abnormal actions.

To address TPAS task, directly transferring mature paradigms from related fields faces fundamental limitations. 
First, traditional Person Re-identification (ReID) and Text-based Person Search (TPS) methods 
highly rely on ``identity and static appearance consistency''~\cite{lyu2025towards,wu2025dagait,long2026towards,li2026mtavg}. They treat pose variations and actions as ``domain noise'' to be eliminated~\cite{lyu2025tempo,lyucome,zhou2026mtavg,long2025enhancing,long2025revisiting,lyu2026hallu_sae}, lacking fine-grained action perception to distinguish between ``a man in red standing'' and ``a man in red falling''.
As shown in Figure~\ref{fig:intro}(a), traditional TIPR models are easily deceived by static appearance, misclassifying normal actions as positive. Conversely, TPAS requires precisely capturing abnormal actions alongside static appearance matching.
Second, traditional Video Anomaly Detection (VAD) methods~\cite{3-N1} extract temporal features for coarse-grained labels or anomaly scores, lacking flexibility for fine-grained retrieval via text query.
Finally, recent TPAS frameworks (e.g., CMP~\cite{CMP2025}) explicitly supplement action features using external human keypoint estimators. This strategy is extremely fragile against severe occlusions, extreme abnormal poses, or low resolutions in real-world surveillance.
Thus, abandoning external detectors and utilizing rich textual semantic priors for action-appearance feature decoupling is essential for solving TPAS. However, this faces the following two severe challenges.

\textbf{C1: Visual Decoupling Failure under Pixel-Level Entanglement}. 
In TPAS, implicit action semantics are highly coupled with explicit static appearance (e.g., limb posture conveys both appearance and action). This pixel-level entanglement makes extracting action features directly in the pure visual space highly unreliable.
Traditional soft decoupling methods~\cite{2-N28} cannot mathematically guarantee feature exclusivity, causing dominant appearance information to  easily contaminate subtle action features.
As shown in Figure~\ref{fig:intro}(b), although action heatmaps initially anchor visual attention to local action regions, the feature response distribution indicates that static appearance still dominates and couples with the action response.
Consequently, even when attention focuses on the action region, extracted features remain unreliably contaminated. Given natural language's fine-grained semantic compositionality (i.e., nouns for appearance, verbs for actions), the primary challenge in TPAS is how to introducing reliable textual semantic priors as anchors to mathematically achieve strict appearance-action decoupling during forward representation.

\textbf{C2: Harmful Shortcut in Hard Negative Optimization}. 
Although geometric constraints (e.g., orthogonal projection) can decouple forward representations, TPAS struggles with hard negatives exhibiting ``same appearance, different action'' (i.e., two people who look exactly the same but different actions). When contrastive losses (e.g., InfoNCE~\cite{Align-before-fuse}) attempt to forcibly separate these visually highly similar samples, they incur significant gradient penalties.
As shown in the Baseline of Figure~\ref{fig:intro}(c), this penalty is intuitively reflected in the action direction's gradient dynamics: the gradient norm ($\|\nabla \mathcal{L}\|_2$) of difficult negative samples increases abnormally, showing a sharp gap compared to positive and simple negative samples.
Furthermore, research has shown~\cite{geirhos2020} that, under extreme optimization pressure, conventional gradient flow in Euclidean space often disregards decoupling geometric constraints, exhibiting severe ``shortcut learning''. Instead of laboriously extracting subtle action features, the optimizer breaks orthogonality and distorts easily optimizable appearance mapping rules in action semantics (e.g., forcing a ``red shirt'' feature to align with ``falling''), leading to manifold drift. Thus, the second key challenge is constraining backpropagation trajectories to eliminate these harmful shortcuts.

To tackle these challenges, we propose the \textbf{Light}weight \textbf{A}ction \textbf{I}nversion and \textbf{R}iemannian rectification network (\textbf{LightAIR}). It comprises three key modules ensuring feature reliability from static representation to dynamic optimization: 
(a) Action Inversion Operator (AIO), which introduces textual semantic priors as anchors and uses a lightweight network to extract reliable action features, overcoming visual-inherent coupling. 
(b) Orthogonal Null-Space Projection (ONSP), which constrains appearance features within orthogonal complement space of action features, avoiding contamination in forward representation. 
(c) Gradient Rectification (GR), which computes Riemannian gradients to constrain backpropagation trajectory. It forces gradient flow to update strictly along tangent space that preserves decoupling properties, cutting off shortcut learning paths and smoothing hard negative gradient differences to converge within a reasonable range (as shown in Figure~\ref{fig:intro}(c), right).
Our contributions are summarized as follows:
\begin{itemize}[leftmargin=8pt]
    \item We deeply analyze the fundamental challenges of cross-modal retrieval in TPAS, revealing for the first time the forward decoupling dilemma caused by pixel-level entanglement, alongside shortcut learning during hard negative optimization.
    \item We propose LightAIR, eliminating reliance on external pose estimators. Through null-space projection and Riemannian gradient rectification, it mathematically achieves robust action-appearance feature decoupling and cross-modal alignment.
    \item Extensive experiments on widely used TPAS datasets (e.g., PAB benchmark) demonstrate that LightAIR significantly outperforms existing state-of-the-art methods across all retrieval metrics.
\end{itemize}

%% file: 2_rel.tex
\section{Related Work}
Our work is closely related to Text-based Person Anomaly Search and Shortcut Learning.

\noindent \textit{\textbf{Text-based Person Anomaly Search.}}
Traditional text-based person retrieval (TIPR) primarily focuses on matching static appearance features~\cite{TIPCB,IRBTD}, whereas conventional video anomaly detection (VAD) emphasizes capturing coarse-grained anomalous actions~\cite{TbVAD,deng2025}.
To address this, Yang et al.~\cite{CMP2025} proposed a new task called Text-based Person Anomaly Search (TPAS). This task requires models to maintain precise appearance matching capabilities while capturing specific anomal person actions at a fine-grained level, thereby enabling accurate retrieval.
There have been several preliminary explorations of this highly challenging emerging task with the development of visual understanding~\cite{zhong2025ctd,zhong2026dyn,wu2026promsa,HABIT,HUD,COMBINER,meng2026clcr,fu2026livegraph,zhao2026resilphase,zhang2026coupling,xu2026psgait} and multimodal learning~\cite{TEMA,fu2026s,luo2026multipress,zhang2026finsentllm,zhu2026ants,zhang2026logical,zhong2026semi}. For example, Yang et al.~\cite{CMP2025} introduced an external human keypoint estimator to explicitly enhance action features and utilized a locally aware alignment mechanism to match text with anomal actions; Ju et al. \cite{AnomalyLMM} leveraged the generative prior of multimodal large models to assist the retrieval process through cross-modal knowledge distillation.
However, existing methods often struggle with feature decoupling and hard negatives when addressing TPAS tasks, making it difficult to meet the stringent requirements of joint action-appearance retrieval~\cite{zhong2024causal,song1,song2,zhong2025adaptive,zhong2024causal}. In contrast, our proposed LightAIR leverages textual semantic priors and Riemannian gradient constraints to effectively ensure robustness in TPAS scenarios.

\begin{figure*}[t!]
	\includegraphics[width=\linewidth]{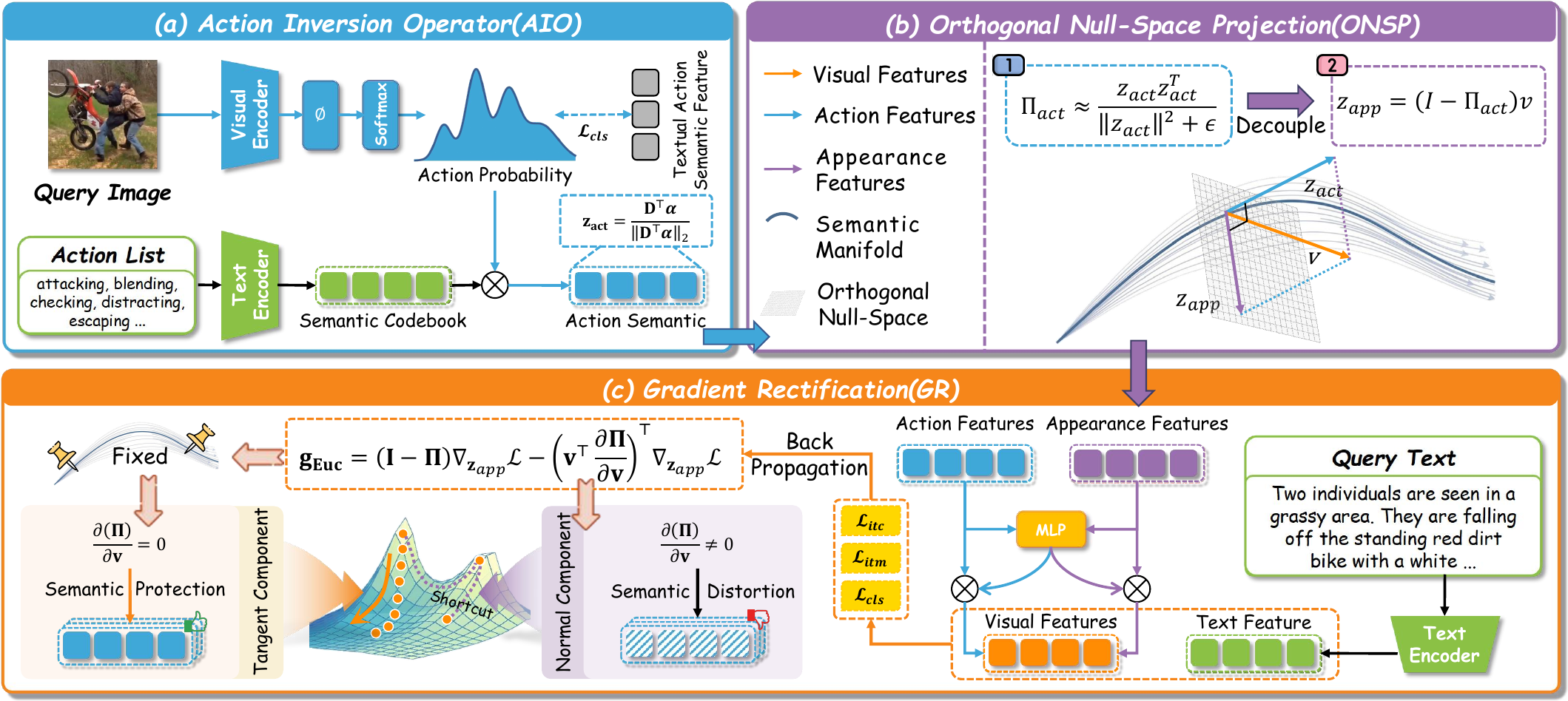}
	\caption{LightAIR consists of \textit{(a) Action Inversion Operator}, which utilizes textual semantic priors to extract pure action features; \textit{(b) Orthogonal Null-Space Projection}, which ensures strict forward decoupling by projecting raw visual features; and \textit{(c) Gradient Rectification}, which computes riemannian gradients to effectively cut off harmful shortcuts.}
	\label{fig:Framework}
\end{figure*}
\noindent
\textit{\textbf{Shortcut Learning.}}
As core challenges to deep learning generalization~\cite{ConeSep,song4,Air-know,song3,yu2025amplifying,shi2026enhancing}, shortcut learning~\cite{yu2026dismantling,TempRet,INTENT,song5} have garnered widespread attention~\cite{Spare, teney2022,yu2025bridging} these years. This has driven extensive research into visual model robustness~\cite{zhu2026dual,teney2022, PDE,song6,maspo,yu2026causally,yu2025bimodal}, cross-modal and generative representation decoupling~\cite{song8,HyGDL,zhou2023causal, Chimera,song7,se_agent,yu2023mixup}, and out-of-distribution (OOD) generalization~\cite{wang2025, Spare,curriculum_rlaif,proxmo}.
Notably, shortcut elimination and feature decoupling mechanisms have recently been extensively explored in multimodal learning~\cite{fu2025adaptive,fu2026cityguard}, including TIPR~\cite{CFAM,TIPS}. For example, Zuo et al.~\cite{PLOT} noted that retrieval models easily fall into coarse-grained textual shortcuts and proposed ultra-fine-grained cross-modal feature mining to mitigate alignment bias. Yu et al.~\cite{CAMeL} forcefully stripped identity-irrelevant visual shortcuts from the latent space via meta-learning adaptation and decoupled conceptual representations, achieving purer semantic isolation. 
However, TPAS task presents a more complex scenario: it requires simultaneously matching action and appearance while preventing their mutual interference from generating harmful shortcuts. Our proposed LightAIR utilizes Riemannian gradient rectification to enforce the correct optimization direction, achieving precise text-based person anomaly search.

%% file: 3_met.tex
\section{LightAIR}

As the major innovation, our proposed LightAIR introduces text semantic priors to extract action anchors, utilizes null-space projection to decouple action features, and employs gradient rectification to maintain semantic stability during optimization. As shown in Figure~\ref{fig:Framework}, LightAIR consists of three key modules: \textit{(a) Action Inversion Operator}, \textit{(b) Orthogonal Null-Space Projection} and \textit{(c) Gradient Rectification}. In this section, we first define the TPAS task and then elaborate on each module.

\subsection{Problem Formulation}

The TPAS task aims to retrieve pedestrian images matching text queries describing appearance and normal or anomaly actions. Formally, let $\mathcal{T}=\{(\mathbf{x}_\mathbb{T},\mathbf{x}_\mathbb{I})_{n}\}_{n=1}^{N}$ be a text and image set of size $N$, where $\mathbf{x}_\mathbb{T}$ and $\mathbf{x}_\mathbb{I}$ denote the text query and pedestrian image respectively. The objective is to jointly optimize the text encoder $\Phi_\mathbb{T}$ and image encoder $\Phi_\mathbb{I}$ to map matching pairs into a shared space, such that $\Phi_\mathbb{T}(\mathbf{x}_\mathbb{T}) \rightarrow \Phi_\mathbb{I}(\mathbf{x}_\mathbb{I})$.

\subsection{Action Inversion Operator (AIO)}
In TPAS, the severe coupling of implicit actions and static appearances makes pure visual action extraction unreliable. To extract robust action representations, we propose the \textit{Action Inversion Operator (AIO)} module (Figure~\ref{fig:Framework}(a)). AIO leverages text semantic priors as anchors and an information bottleneck-based reconstruction mechanism to achieve deep action-appearance decoupling.

\noindent \textbf{Global Semantic Codebook Construction.}
To introduce text semantic priors as precise action anchors, we first automatically extract reliable action words from the training dataset. Using NLTK~\cite{NLTK} and SpaCy~\cite{SpaCy}, we perform part-of-speech tagging and dependency parsing on full-sentence descriptions. By explicitly filtering out redundant words related to entity appearances and background scenes, we construct a representative action set $\mathcal{A} = \{a_1, a_2, \dots, a_L\}$ of size $L$.

During initialization, we extract semantic features from this set using a pre-trained text encoder $\Phi_{text}$. Normalizing these features yields the action semantic anchors, formulated as,
\begin{equation}
    \mathbf{d}_l = \frac{\Phi_{text}(a_L)}{\|\Phi_{text}(a_L)\|_2}.
\end{equation}
Subsequently, we stack these semantic feature vectors row-wise to construct a globally shared semantic codebook $\mathbf{D} \in \mathbb{R}^{L \times D}$ as,
\begin{equation}
    \mathbf{D} = [\mathbf{d}_1, \mathbf{d}_2, \dots, \mathbf{d}_L]^\top,
\end{equation}
where $L$ denotes the number of action semantic anchors and $D$ represents the feature dimension. Furthermore, to ensure anchor stability, this semantic codebook is frozen after initialization and excluded from subsequent network parameter updates.

\noindent \textbf{Latent Coefficient Estimation.} 
For a given pedestrian image $\mathbf{x}_\mathbb{I}$, the global feature $\mathbf{v} \in \mathbb{R}^{D}$ extracted by the pre-trained visual encoder exhibits a high coupling between static appearance and dynamic action. To isolate the pure action feature from the entangled visual space, we project $\mathbf{v}$ into a reliable semantic subspace spanned by the global codebook $\mathbf{D}$.

To prevent appearance noise from direct visual-codebook interactions, we design a lightweight \textit{Latent Coefficient Estimator} $\Phi_{act}$. Serving as an information bottleneck, it implicitly maps $\mathbf{v}$ to subspace coordinate coefficients and applies Top-K sparsification to suppress redundant action semantics, formulated as,

\begin{equation}
    \boldsymbol{\alpha} = \text{Softmax}(\text{Top-K}(\Phi_{act}(\mathbf{v}))),
    \label{eq:topk}
\end{equation}
where $\text{Top-K}(\cdot)$ retains only the top $K$ activations while masking the rest, and $\boldsymbol{\alpha} \in \mathbb{R}^{L}$ denotes the sparse activation coefficient across $L$ action semantic bases. This compels the model to describe complex actions using only a few essential action components.

\noindent \textbf{Codebook-driven Reconstruction \& Alignment.} 
After obtaining the activation coefficient $\boldsymbol{\alpha}$, we physically reconstruct the action visual feature using the frozen reliable action semantic codebook $\mathbf{D}$ as a basis. Specifically, the reconstructed action visual feature $\mathbf{z}_{act}\in \mathbb{R}^{D}$ is obtained via a weighted linear combination of the codebook and $\boldsymbol{\alpha}$, formulated as,
\begin{equation}
    \mathbf{z}_{act} = \frac{\mathbf{D}^\top \boldsymbol{\alpha}}{\|\mathbf{D}^\top \boldsymbol{\alpha}\|_2}.
    \label{eq:zact}
\end{equation}
Architecturally, $\Phi_{act}$ and $\mathbf{D}$ form a structured encoder-decoder process where $\Phi_{act}$ estimates relevant action coefficients from coupled visual semantics, and $\mathbf{D}$ acts as a fixed basis for decoding.

Since forward latent estimation lacks explicit text guidance, accurately mapping subtle visual actions to the semantic bases in $\mathbf{D}$ remains challenging. To enhance the cross-modal mapping and reconstruction precision of $\Phi_{act}$, we introduce text-guided semantic estimation alignment. Specifically, we extract ground-truth action feature $\mathbf{t}_y \in \mathbb{R}^{D}$ from the text query $\mathbf{x}_\mathbb{T}$ via $\Phi_\mathbb{T}$, and align the reconstructed $\mathbf{z}_{act}$ with $\mathbf{t}_y$ via a discriminative loss $\mathcal{L}_{cls}$ as,

\begin{equation}
    \mathcal{L}_{cls} = 1-Cosine(\mathbf{z}_{act}, \mathbf{t}_y),
    \label{eq:cls}
\end{equation}
This explicit supervision constrains $\Phi_{act}$'s optimization via backpropagation. Ultimately, this closed loop of implicit estimation, frozen reconstruction, and semantic calibration improves mapping accuracy of $\Phi_{act}$, yielding reliable action features.

\subsection{Orthogonal Null-Space Projection (ONSP)}
Alongside extracting reliable action features $\mathbf{z}_{act}$, TPAS requires accurate static appearance extraction. Since traditional loss-based soft decoupling lacks explicit geometric constraints~\cite{locatello2019}, it often leaves dynamic action leakage within appearance features. Thus, inspired by feature debiasing~\cite{INLP}, we propose the \textit{Orthogonal Null-Space Projection} module (Figure~\ref{fig:Framework}(b)). Abandoning black-box extractions, this module utilizes null-space projection to strictly decouple appearance and action at the geometric level.

\noindent \textbf{Action Subspace Construction.} 

To isolate action information from the original visual feature, we first define the action subspace within the multimodal metric space. Reconstructed via the text semantic codebook, the action feature $\mathbf{z}_{act}$ indicates the action direction and serves as the natural basis for this subspace. Accordingly, we construct an orthogonal projection operator $\mathbf{\Pi}_{act} \in \mathbb{R}^{D \times D}$ to extract the visual feature component aligned with the current action semantics, formulated as,
\begin{equation}
    \mathbf{\Pi}_{act} = \mathbf{z}_{act} (\mathbf{z}_{act}^\top \mathbf{z}_{act} + \epsilon)^{-1} \mathbf{z}_{act}^\top \approx \frac{\mathbf{z}_{act} \mathbf{z}_{act}^\top}{\|\mathbf{z}_{act}\|^2 + \epsilon},
\label{eq:pi}
\end{equation}
where $\mathbf{z}_{act}$ is the action feature derived from Eq.~\ref{eq:zact}, and $\epsilon$ is a minimal constant preventing zero division.

\noindent \textbf{Nullspace Mapping of Appearance Features.} 
Since the previously calculated projection operator $\mathbf{\Pi}_{act}$ fully captures the action semantic direction in the feature space, its orthogonal complement space $\mathcal{S}_{act}^{\perp}$ naturally excludes action information. Thus, we project the original global visual feature $\mathbf{v}$ into the null-space of $\mathbf{\Pi}_{act}$. Through this orthogonal projection, we mathematically eliminate the coupled action component in $\mathbf{v}$ to obtain the pure static appearance feature $\mathbf{z}_{app}\in \mathbb{R}^{D}$, formulated as,
\begin{equation}
    \mathbf{z}_{app} = (\mathbf{I} - \mathbf{\Pi}_{act})\mathbf{v},
\label{eq:nullspace}
\end{equation}
where $\mathbf{I} \in \mathbb{R}^{D \times D}$ is the identity matrix.

\subsection{Gradient Rectification (GR)}
Since action features are inherently weaker than appearance features, the model often exploits harmful optimization shortcuts during backpropagation to separate hard negatives~\cite{geirhos2020}. This misattributes appearance variances to action differences, causing severe overfitting. To resolve this, we propose \textit{Gradient Rectification} module (Figure~\ref{fig:Framework}(c)), which severs these shortcuts at the gradient level, confining update trajectories within the valid semantic subspace.

\noindent \textbf{Preliminary: Diagnosing Shortcut Learning via Gradient Decomposition.} 
First, we mathematically analyze how backpropagation distorts the feature decoupling mechanism. Since obtaining the pure appearance feature $\mathbf{z}_{app}$ relies on the action projection operator $\mathbf{\Pi}_{act}$ (Eq.~\ref{eq:nullspace}), we apply the total differential chain rule to decompose the global loss gradient with respect to the visual feature $\mathbf{v}$. Specifically, for the appearance branch, the backpropagated Euclidean gradient $\mathbf{g}_{Euc} \in \mathbb{R}^{D}$ is decomposed as,

\begin{equation}
\small
\begin{aligned}
        \mathbf{g}_{Euc} &= \frac{\partial \mathcal{L}}{\partial \mathbf{z}_{app}} \cdot \frac{d \mathbf{z}_{app}}{d \mathbf{v}}
        = \left( \frac{\partial \mathcal{L}}{\partial \mathbf{z}_{app}} \right)^T \left[ (\mathbf{I} - \mathbf{\Pi}) - \left( \frac{\partial \mathbf{\Pi}}{\partial \mathbf{v}} \times \mathbf{v} \right) \right]\\
        &= \underbrace{(\mathbf{I} - \mathbf{\Pi})\nabla_{\mathbf{z}_{app}}\mathcal{L}}_{\text{Tangent Component}} - \underbrace{\left( \frac{\partial\mathbf{\Pi}}{\partial\mathbf{v}} \times_1 \mathbf{v} \right)^\top \nabla_{\mathbf{z}_{app}}\mathcal{L}}_{\text{Normal Component}}
\end{aligned}
\label{eq:Euc}
\end{equation}
where $\mathcal{L}$ is the global loss, and $\times_1$ denotes tensor-vector contraction. Eq.~\ref{eq:Euc} reveals that the visual feature's gradient decomposes into two orthogonal components,

\begin{itemize}[leftmargin=8pt]
   \item \textbf{Tangent Component}: The valid optimization direction. It restricts appearance updates within the action-free null-space to safely increase sample distances.
   \item \textbf{Normal Component}: It contains the partial derivative $\frac{\partial\mathbf{\Pi}}{\partial\mathbf{v}}$. Since deriving $\mathbf{z}_{app}$ relies on $\mathbf{\Pi}$ (Eq.~\ref{eq:nullspace}), the dominant appearance gradient $\nabla_{\mathbf{z}_{app}}\mathcal{L}$ oversteps during backpropagation, improperly attempting to alter $\mathbf{\Pi}$'s underlying mapping rules.
\end{itemize}

When confronting hard negatives, the \textit{Normal Component} generates massive gradients to rapidly minimize loss, triggering optimization shortcuts. Instead of learning discriminative action features, the optimizer exploits these overstepping appearance gradients to distort the action extraction mechanism (e.g., misclassifying clothing as an action). This cross-branch contamination fundamentally causes the action semantic manifold drift.

\noindent \textbf{Riemannian Gradient Projection.} 
To eliminate these cross-branch harmful shortcuts, we rectify the Euclidean gradient, restricting visual feature updates to the Tangent Space $T_{\mathbf{v}}\mathcal{M}$ of the action semantic manifold $\mathcal{M}$ at $\mathbf{v}$.

Under this geometric constraint, the steepest descent on the action manifold $\mathcal{M}$ is defined as the Riemannian gradient~\footnote{The Riemannian gradient equates to the orthogonal projection of the Euclidean gradient onto the local tangent space.} $\mathbf{g}_{Riem} \in \mathbb{R}^{D}$~\cite{absil2008}, obtained by orthogonally projecting the Euclidean gradient onto the local tangent space, formulated as,
\begin{equation}
    \mathbf{g}_{Riem} = \text{Proj}_{T_{\mathbf{v}}\mathcal{M}}(\mathbf{g}_{Euc}) = (\mathbf{I} - \mathbf{\Pi}_{act})\mathbf{g}_{Euc},
\label{eq:riemannian projection}
\end{equation}
where $T{\mathbf{v}}\mathcal{M} = \{ \boldsymbol{\xi} \in \mathbb{R}^D \mid \mathbf{\Pi}_{act} \boldsymbol{\xi} = \mathbf{0} \}$ is the null-space of $\mathbf{\Pi}_{act}$, encompassing all valid change directions that preserve the action semantic basis, while $\text{Proj}_{T_{\mathbf{v}}\mathcal{M}}$ denotes orthogonal projection onto this tangent space.

\noindent \textbf{Entropy-Guided Retraction.} 
Although the Riemannian gradient provides the correct theoretical optimization direction, strictly enforcing the Riemannian gradient early in training (when $\Phi_{act}$ remains unreliable) can sever beneficial explorations and cause optimization stagnation. To smoothly transition optimization, we implement an adaptive weighting mechanism based on uncertainty estimation. Specifically, we quantify the model's mapping confidence via the Shannon entropy of the action activation probability,
\begin{equation}
    \mathcal{H} = - \sum \alpha_l \log \alpha_l, \quad \omega(\mathbf{v}) = \exp(-\mathcal{H}/\tau)
\label{eq:omega}
\end{equation}
where $\alpha_l$ is the probability activation vector of the $l$-th action semantic anchor, $\mathcal{H}$ is the entropy of the action activation probability distribution, and $\omega \in [0,1]$ is the uncertainty weight coefficient.

To balance early fast convergence and late semantic stability, we dynamically combine the original Euclidean and rectified Riemannian gradients. The final rectified gradient $\mathbf{g}_{final} \in \mathbb{R}^{D}$ for the visual feature is formulated as,
\begin{equation}
\begin{aligned}
    \mathbf{g}_{final} &=(1 - \omega)\mathbf{g}_{Euc} + \omega \mathbf{g}_{Riem} \\
    &= (1 - \omega) \mathbf{g}_{Euc} + \omega \text{Proj}_{T_{\mathbf{v}}\mathcal{M}} (\mathbf{g}_{Euc}) 
    = \mathbf{g}_{Euc} - \omega \mathbf{\Pi} \mathbf{g}_{Euc}
\end{aligned}
\label{eq:g_final}
\end{equation}
During early training with high entropy ($\omega \rightarrow 0$), the model uses the Euclidean gradient to accelerate optimization. As confidence in action semantics increases ($\omega \rightarrow 1$), it shifts to the Riemannian gradient to avoid harmful shortcuts.

\noindent \textbf{Adaptive Semantic Aggregation and Overall Objective.}
After determining the correct gradient direction, we combine $\mathbf{z}_{app}$ and $\mathbf{z}_{act}$ for cross-modal matching with text queries. While linear addition facilitates independent gradient shunting during backpropagation, it induces ``secondary coupling'' in the forward representation. Specially, dominant appearance information overshadows weaker action signals during similarity computation.
Thus, to ensure independent shunting of backpropagated gradients while avoiding secondary coupling in the forward representation, we propose adaptive semantic aggregation. Specially, it concatenates these orthogonal features along the channel dimension and uses a lightweight MLP to learn adaptive gating weights,
\begin{equation}
    \mathbf{W} = \text{MLP}([\mathbf{z}_{act},\mathbf{z}_{app}]),
\label{eq:mlp}
\end{equation}
where $\mathbf{W}\in\mathbb{R}^{2D}$. Subsequently, weights $\mathbf{W}_{act} \in \mathbb{R}^{D}$ and $\mathbf{W}_{app} \in \mathbb{R}^{D}$ are derived from $\mathbf{W}$ to compute a weighted sum. This dynamically calibrates matching contributions from dual-stream semantics, yielding the final global visual feature $\mathbf{z}_{final}$ for cross-modal retrieval, formulated as,
\begin{equation}
    \mathbf{z}_{final} = \mathbf{W}_{act} \odot \mathbf{z}_{act} + \mathbf{W}_{app} \odot \mathbf{z}_{app}.
\label{eq:combine}
\end{equation}

Gated aggregation prevents appearance interference in the forward pass and shunts backpropagated gradients into independent $\nabla_{\mathbf{z}_{act}}\mathcal{L}$ and $\nabla_{\mathbf{z}_{app}}\mathcal{L}$ paths. This structured shunting establishes the mathematical prerequisite for gradient rectification in Eq.~\ref{eq:g_final}.

Finally, the pre-trained text encoder extracts query global semantic features $\mathbf{t}_q \in \mathbb{R}^D$. For fine-grained image-text alignment, we adopt standard image-text contrastive ($\mathcal{L}_{itc}$) and matching ($\mathcal{L}_{itm}$) losses~\cite{Align-before-fuse}. In a batch of size $B$, we define the temperature-scaled similarity between the $i$-th text query and $j$-th visual feature as $s_{i,j} = \langle \mathbf{t}_{q}^i, \mathbf{z}_{final}^j \rangle / \tau_c$, where $\langle \cdot, \cdot \rangle$ denotes cosine similarity and $\tau_c$ is a learnable coefficient. $\mathcal{L}_{itc}$ then utilizes a symmetric InfoNCE format to calculate bidirectional contrastive losses:

\begin{equation}
    \mathcal{L}_{itc} = - \frac{1}{2B} \sum_{i=1}^{B} \left( \log \frac{\exp(s_{i,i})}{\sum_{j=1}^B \exp(s_{i,j})} + \log \frac{\exp(s_{i,i})}{\sum_{j=1}^B \exp(s_{j,i})} \right).
\label{eq:itc}
\end{equation}
Alternatively, $\mathcal{L}_{itm}$ utilizes binary classification head $\Phi_{itm}$ to predict matching for fused image-text features. During feature aggregation, we follow the sequence of ``text query - visual target'' to perform feature concatenation and calculate the binary cross-entropy:

\begin{equation}
    \mathcal{L}_{itm} = - \frac{1}{B} \sum_{i=1}^{B} \Big( y_i \log p_m^i + (1 - y_i) \log (1 - p_m^i) \Big),
\label{eq:itm}
\end{equation}
where $p_m^i = \Phi_{itm}([\mathbf{t}_q^i \parallel \mathbf{z}_{final}^i])$ denotes the predicted matching probability for the $i$-th pair of image queries conditioned on text, and $y_i \in \{0,1\}$ is the ground-truth matching label. 

Incorporating the action discriminative loss $\mathcal{L}_{cls}$ from Eq.~\ref{eq:cls}, the global optimization objective for LightAIR is formulated as,

\begin{equation}
    \mathbf{\Theta^{*}}=
    \underset{\mathbf{\Theta}}{\arg \min } \left( \mathcal{L}_{itc} + \gamma_1 \mathcal{L}_{itm} + \gamma_2 \mathcal{L}_{cls}\right),
\label{optimization}
\end{equation}
where $\mathbf{\Theta^{*}}$ is the to-be-optimized parameter for LightAIR and $\gamma$ is the trade-off hyper-parameter. Following previous works~\cite{CMP2025,Bi-IRRA}, we set the parameters for $\mathcal{L}_{itc}$ to $1$.

%% file: 4_exp.tex
\section{Experiment}
In this section, we first outline the experimental settings, followed by a detailed presentation of the experimental results and corresponding analyses.

\subsection{Experimental Settings}

\subsubsection{Datasets.}

To verify the effectiveness and generalization of the proposed method, we evaluate it on two core tasks: Text-based Person Anomaly Search (TPAS) and Text-to-Image Person Retrieval (TIPR). For the TPAS task, we adopt the fine-grained normal and anomaly action retrieval dataset PAB~\cite{PAB} (Pedestrian Anomaly Behavior) and introduce the MultiWeather setting containing 10 simulated weather conditions alongside the out of distribution (OOD) test set UCC to assess model robustness under extreme conditions. For the TIPR task, we select the CUHK-PEDES~\cite{CUHK-PEDES}, ICFG-PEDES~\cite{ICFG-PEDES}, and RSTPReid~\cite{RSTPReID} datasets, along with the UFineBench~\cite{CFAM} dataset which features open domain and ultra fine grained text descriptions.

\begin{table*}[t!]
  \centering
  \caption{Performance comparison on the PAB dataset of multi-weather test setting, measured by R@1 and mAP for each weather. The overall best results are indicated in \textbf{bold}.}
  \vspace{-12pt}
    \resizebox{\linewidth}{!}{%
    \begin{tabular}{c|cc|cc|cc|cc|cc|cc|cc|cc|cc|cc|cc}
 \Xhline{1.5pt}
    \multirow{2}{*}{Method} & \multicolumn{2}{c|}{Normal} & \multicolumn{2}{c|}{Wind} & \multicolumn{2}{c|}{Rain} & \multicolumn{2}{c|}{Snow} & \multicolumn{2}{c|}{Rain+Snow} & \multicolumn{2}{c|}{Dark} & \multicolumn{2}{c|}{Dark+Wind} & \multicolumn{2}{c|}{Dark+Rain} & \multicolumn{2}{c|}{Dark+Snow} & \multicolumn{2}{c|}{Over-exposure} & \multicolumn{2}{c}{Mean} \\
    \multicolumn{1}{c|}{} & \multicolumn{1}{c}{R@1} & \multicolumn{1}{c|}{mAP} & \multicolumn{1}{c}{R@1} & \multicolumn{1}{c|}{mAP} & \multicolumn{1}{c}{R@1} & \multicolumn{1}{c|}{mAP} & \multicolumn{1}{c}{R@1} & \multicolumn{1}{c|}{mAP} & \multicolumn{1}{c}{R@1} & \multicolumn{1}{c|}{mAP} & \multicolumn{1}{c}{R@1} & \multicolumn{1}{c|}{mAP} & \multicolumn{1}{c}{R@1} & \multicolumn{1}{c|}{mAP} & \multicolumn{1}{c}{R@1} & \multicolumn{1}{c|}{mAP} & \multicolumn{1}{c}{R@1} & \multicolumn{1}{c|}{mAP} & \multicolumn{1}{c}{R@1} & \multicolumn{1}{c|}{mAP} & \multicolumn{1}{c}{R@1} & \multicolumn{1}{c}{mAP} \\
    \hline
    \multicolumn{1}{l|}{X-VLM~\cite{X-VLM}~\textcolor{gray}{(ICML'22)}} & 83.47  & 90.94  & 79.02  & 88.10  & 54.40  & 67.40  & 59.10  & 72.34  & 49.95  & 62.09  & 79.58  & 88.24  & 75.53  & 85.47  & 34.93  & 45.98  & 50.25  & 63.32  & 74.87  & 84.82  & 64.11  & 74.87  \\
    \multicolumn{1}{l|}{CMP~\cite{CMP2025}~\textcolor{gray}{(ICCV'25)}}   & 84.93  & 91.66  & 81.24  & 89.34  & 60.06  & 72.53  & 63.40  & 75.74  & 54.85  & 67.31  & 80.89  & 89.00  & 77.20  & 86.55  & 39.03  & 50.58  & 53.49  & 66.12  & 76.09  & 85.56  & 67.12  & 77.44  \\
    \hline
    \rowcolor[rgb]{0.910, 0.941, 0.980}  \multicolumn{1}{l|}{\textbf{LightAIR~(Ours)}}  &\textbf{85.49} &\textbf{92.20} &\textbf{81.43} &\textbf{90.63} &\textbf{61.76} &\textbf{74.34} &\textbf{65.47} &\textbf{78.12} &\textbf{58.63} &\textbf{73.44} &\textbf{81.37} &\textbf{90.08} &\textbf{79.19} &\textbf{88.85} &\textbf{41.31} &\textbf{52.58} &\textbf{55.91} &\textbf{68.49} &\textbf{78.15} &\textbf{88.14} &\textbf{68.87} &\textbf{79.69}\\
 \Xhline{1.5pt}
    \end{tabular}%
    }
  \vspace{-13pt}
  \label{tab:multi}%
\end{table*}%

\subsubsection{Implementation Details.}

Following previous work~\cite{PAB}, the training batch size of LightAIR is set to $22$. We initialize the encoder weights using X2VLM~\cite{x2vlm} and employ the AdamW optimizer with a weight decay of $0.01$. The learning rate is initialized as $2e-5$. Number $K$ in Eq.~\ref{eq:topk} is set to $3$. 
The image text matching loss weight $\gamma_1$ and the action discriminative loss weight $\gamma_2$ are set to $4$ and $1$, respectively. We equivalently implement Riemannian gradient rectification by applying a gradient clipping operation to $\mathbf{\Pi}_{act}$. All experiments were conducted on a single NVIDIA V100 GPU with $32$ GB memory and trained for $20$ epochs.

\subsubsection{Evaluation.} 

To ensure fair comparisons, we follow the standard evaluation protocols of each dataset, adopting Recall@k (R@k) and mean Average Precision (mAP) as evaluation metrics. For the TPAS task, we report R@\{1, 5, 10\} and mAP on the PAB and UCC datasets; in the Multi-weather evaluation, R@1 and mAP are compared. For the TIPR task, we report the mean R@\{1, 5, 10\} across the CUHK-PEDES, ICFG-PEDES, RSTPReid, and UFineBench datasets. Specifically, on the UFineBench dataset, results under both UFine6926 and UFine3C settings are provided.

\subsection{Performance Comparison}

To systematically verify the performance and generalization of LightAIR, we conduct a series of performance evaluation experiments on the TPAS and TIPR tasks and compare it with existing SOTA models.

\subsubsection{On TPAS Task.}

\begin{table}[h]
  \centering
  \small
  \vspace{-10pt}
  \tabcolsep=10pt
  \caption{Performance comparison on the PAB dataset, measured by R@K and mAP. Over-all best results are in bold, and the sub-optimal is underlined.}
  \vspace{-10pt}
    \resizebox{0.98\linewidth}{!}{%
    \begin{tabular}{rrrrrr}
 \Xhline{1.5pt}
    \multicolumn{1}{c|}{Method} & \multicolumn{1}{c|}{\#Data} & \multicolumn{1}{c}{R@1} & \multicolumn{1}{c}{R@5} & \multicolumn{1}{c}{R@10} & \multicolumn{1}{c}{mAP} \\
    \hline
    \multicolumn{1}{l|}{CLIP~\cite{clip}~\textcolor{gray}{(ICML'21)}} & \multicolumn{1}{c|}{-} & \multicolumn{1}{c}{47.57} & \multicolumn{1}{c}{81.55} & \multicolumn{1}{c}{89.03} & \multicolumn{1}{c}{62.73} \\
    \multicolumn{1}{l|}{X-VLM~\cite{X-VLM}~\textcolor{gray}{(ICML'22)}} & \multicolumn{1}{c|}{-} & \multicolumn{1}{c}{71.94} & \multicolumn{1}{c}{97.78} & \multicolumn{1}{c}{98.99} & \multicolumn{1}{c}{83.96} \\
    \multicolumn{1}{l|}{RaSa~\cite{RaSa}~\textcolor{gray}{(IJCAI'23)}} & \multicolumn{1}{c|}{-} & \multicolumn{1}{c}{21.74} & \multicolumn{1}{c}{27.30} & \multicolumn{1}{c}{27.96} & \multicolumn{1}{c}{24.35} \\
    \multicolumn{1}{l|}{APTM~\cite{APTM}~\textcolor{gray}{(MM'23)}} & \multicolumn{1}{c|}{-} & \multicolumn{1}{c}{22.90} & \multicolumn{1}{c}{45.80} & \multicolumn{1}{c}{52.38} & \multicolumn{1}{c}{33.56} \\
    \multicolumn{1}{l|}{IRRA~\cite{IRRA}~\textcolor{gray}{(CVPR'23)}} & \multicolumn{1}{c|}{-} & \multicolumn{1}{c}{30.59} & \multicolumn{1}{c}{59.61} & \multicolumn{1}{c}{68.91} & \multicolumn{1}{c}{44.41} \\ 
    \multicolumn{1}{l|}{MRA~\cite{MRA}~\textcolor{gray}{(arXiv'25)}} & \multicolumn{1}{c|}{-} & \multicolumn{1}{c}{9.91} & \multicolumn{1}{c}{23.66} & \multicolumn{1}{c}{31.45} & \multicolumn{1}{c}{17.15} \\    
    \multicolumn{1}{l|}{WoRA~\cite{WoRA}~\textcolor{gray}{(WWW'25)}} & \multicolumn{1}{c|}{-} & \multicolumn{1}{c}{22.25} & \multicolumn{1}{c}{45.91} & \multicolumn{1}{c}{53.54} & \multicolumn{1}{c}{33.39} \\    
    \multicolumn{1}{l|}{CAMeL~\cite{CAMeL}~\textcolor{gray}{(TIFS'25)}} & \multicolumn{1}{c|}{-} & \multicolumn{1}{c}{24.47} & \multicolumn{1}{c}{50.00} & \multicolumn{1}{c}{58.75} & \multicolumn{1}{c}{36.75} \\
    \hline
    \multicolumn{1}{l|}{CLIP~\cite{clip}~\textcolor{gray}{(ICML'21)}} & \multicolumn{1}{c|}{0.1M} & \multicolumn{1}{c}{77.60} & \multicolumn{1}{c}{98.84} & \multicolumn{1}{c}{99.75} & \multicolumn{1}{c}{87.35} \\
    \multicolumn{1}{l|}{X-VLM~\cite{X-VLM}~\textcolor{gray}{(ICML'22)}} & \multicolumn{1}{c|}{0.1M} & \multicolumn{1}{c}{81.95} & \multicolumn{1}{c}{98.84} & \multicolumn{1}{c}{99.19} & \multicolumn{1}{c}{89.86} \\
    \multicolumn{1}{l|}{RaSa~\cite{RaSa}~\textcolor{gray}{(IJCAI'23)}} & \multicolumn{1}{c|}{0.1M} & \multicolumn{1}{c}{80.79} & \multicolumn{1}{c}{98.89} & \multicolumn{1}{c}{99.65} & \multicolumn{1}{c}{89.20} \\
    \multicolumn{1}{l|}{APTM~\cite{APTM}~\textcolor{gray}{(MM'23)}} & \multicolumn{1}{c|}{0.1M} & \multicolumn{1}{c}{72.14} & \multicolumn{1}{c}{95.30} & \multicolumn{1}{c}{97.17} & \multicolumn{1}{c}{82.78} \\
    \multicolumn{1}{l|}{IRRA~\cite{IRRA}~\textcolor{gray}{(CVPR'23)}} & \multicolumn{1}{c|}{0.1M} & \multicolumn{1}{c}{76.39} & \multicolumn{1}{c}{97.62} & \multicolumn{1}{c}{99.14} & \multicolumn{1}{c}{86.33} \\
    \multicolumn{1}{l|}{MRA~\cite{MRA}~\textcolor{gray}{(arXiv'25)}} & \multicolumn{1}{c|}{0.1M} & \multicolumn{1}{c}{70.53 } & \multicolumn{1}{c}{94.69} & \multicolumn{1}{c}{97.47} & \multicolumn{1}{c}{81.59} \\     
    \multicolumn{1}{l|}{WoRA~\cite{WoRA}~\textcolor{gray}{(WWW'25)}} & \multicolumn{1}{c|}{0.1M} & \multicolumn{1}{c}{74.47 } & \multicolumn{1}{c}{96.82} & \multicolumn{1}{c}{98.48} & \multicolumn{1}{c}{84.60} \\
    \multicolumn{1}{l|}{CAMeL~\cite{CAMeL}~\textcolor{gray}{(TIFS'25)}} & \multicolumn{1}{c|}{0.1M} & \multicolumn{1}{c}{74.30} & \multicolumn{1}{c}{96.79} & \multicolumn{1}{c}{98.84} & \multicolumn{1}{c}{84.20} \\
    \multicolumn{1}{l|}{CMP~\cite{CMP2025}~\textcolor{gray}{(ICCV'25)}} & \multicolumn{1}{c|}{0.1M} & \multicolumn{1}{c}{83.06 } & \multicolumn{1}{c}{98.89} & \multicolumn{1}{c}{99.49} & \multicolumn{1}{c}{90.41} \\
    \rowcolor[rgb]{0.910, 0.941, 0.980} \multicolumn{1}{l|}{\textbf{LightAIR (Ours)}} & \multicolumn{1}{c|}{0.1M} & \multicolumn{1}{c}{84.73}& \multicolumn{1}{c}{\textbf{99.65}} & \multicolumn{1}{c}{\underline{99.85}}&  \multicolumn{1}{c}{\underline{91.93}} \\
    \hline
    \multicolumn{1}{l|}{CMP~\cite{CMP2025}~\textcolor{gray}{(ICCV'25)}} & \multicolumn{1}{c|}{1M} & \multicolumn{1}{c}{\underline{84.93}}  & \multicolumn{1}{c}{\underline{99.09}} & \multicolumn{1}{c}{99.75} & \multicolumn{1}{c}{91.66} \\
    \rowcolor[rgb]{0.910, 0.941, 0.980}  \multicolumn{1}{l|}{\textbf{LightAIR (Ours)}} & \multicolumn{1}{c|}{1M} &  \multicolumn{1}{c}{\textbf{85.49}} & \multicolumn{1}{c}{\textbf{99.65}} & \multicolumn{1}{c}{\textbf{99.99}} & \multicolumn{1}{c}{\textbf{92.20}} \\
 \Xhline{1.5pt}
          &       &       &       &       &  \\
    \end{tabular}%
    }
    \vspace{-20pt}
    \label{tab:pab}%
\end{table}%

\noindent \textbf{Quantitative Results on TPAS Task.} 

Table~\ref{tab:pab} reports the evaluation results of LightAIR on the PAB dataset. We have the following observations: 1) Existing TIPR models struggle to capture the fine-grained anomaly action required by the TPAS task, facing performance bottlenecks. For example, models designed for the TIPR task, such as RaSa~\cite{RaSa} and IRRA~\cite{IRRA}, experience significant performance drops when directly generalized to the TPAS scenario. 2) LightAIR achieves the best performance across all metrics. At the 0.1M data scale, its R@1 (84.73\%) and mAP (91.93\%) improve upon the second best method CMP by 1.67\% and 1.52\%, respectively, and this advantage is maintained at the 1M scale. This is attributed to LightAIR reconstructing behavior signals utilizing semantic anchors and avoiding harmful shortcuts during model training through the gradient rectification mechanism, providing a clear decision boundary for distinguishing negative samples.

\noindent \textbf{Multi-weather evaluation.} 
Table~\ref{tab:multi} shows that LightAIR consistently outperforms CMP across all ten weather conditions, achieving average R@1 and mAP scores of 68.87\% and 79.69\%, respectively. This corresponds to gains of 1.75 and 2.25 points over CMP; even under Dark+Rain, LightAIR improves R@1 by 2.28 points.

\begin{table}[h]
  \centering
  \tabcolsep=12pt
    \small
  \caption{Comparison with existing methods in the OOD setting, measured by R@k and mAP. The overall best results are indicated in bold, and second-best results are underlined.}
  \vspace{-12pt}
    \resizebox{0.98\linewidth}{!}{%
    \begin{tabular}{c|cccc}
 \Xhline{1.5pt}
    Method & \multicolumn{1}{c}{R@1} & \multicolumn{1}{c}{R@5} & \multicolumn{1}{c}{R@10} & \multicolumn{1}{c}{mAP} \\
    \hline
    \multicolumn{1}{l|}{CLIP~\cite{clip}~\textcolor{gray}{(ICML'21)}}  & 51.60  & 68.31  & 76.43  & 43.05  \\
    \multicolumn{1}{l|}{X-VLM~\cite{X-VLM}~\textcolor{gray}{(ICML'22)}} & 52.33  & 66.73  & 72.54  & 40.87  \\
    \multicolumn{1}{l|}{APTM~\cite{APTM}~\textcolor{gray}{(MM'23)}}  & 27.86  & 40.41  & 46.77  & 22.61  \\
    \multicolumn{1}{l|}{IRRA~\cite{IRRA}~\textcolor{gray}{(CVPR'23)}   }& 40.28  & 57.24  & 65.98  & 33.53  \\
    \multicolumn{1}{l|}{RaSa~\cite{RaSa}~\textcolor{gray}{(IJCAI'23)}  }& 54.12  & 70.32  & 75.96  & 39.71  \\
    \multicolumn{1}{l|}{CMP~\cite{CMP2025}~\textcolor{gray}{(ICCV'25)}}   & 54.12  & 71.07  & 77.90  & 43.13  \\
    \multicolumn{1}{l|}{CMP(1M)~\cite{CMP2025}~\textcolor{gray}{(ICCV'25)}} & 55.23  & 71.67  & 77.99  & 44.35  \\
    \hline
    \rowcolor[rgb]{0.910, 0.941, 0.980} \multicolumn{1}{l|}{\textbf{LightAIR (Ours)}}  & \underline{62.36}  & \underline{78.14}  & 	\underline{86.33}  & \underline{51.53}  \\
    \rowcolor[rgb]{0.910, 0.941, 0.980}  \multicolumn{1}{l|}{\textbf{LightAIR~(1M) (Ours)}} & \textbf{63.27}  & \textbf{78.63}  & \textbf{86.76}  & \textbf{52.25}  \\
 \Xhline{1.5pt}
    \end{tabular}%
    }
  \vspace{-8pt}
  \label{tab:ood}%
\end{table}%

\noindent \textbf{OOD evaluation.} 

Furthermore, we evaluated the generalization of LightAIR on the UCC (OOD) test set. As shown in Table~\ref{tab:ood}, LightAIR achieves the best performance. Notably, using only 0.1M training data, its R@1 (62.36\%) and mAP (51.53\%) surpass the second best method CMP trained on full data. This advantage is attributed to the framework of LightAIR utilizing Riemannian gradient rectification, which accurately selects the correct optimization path on cross-domain data rather than merely overfitting to the training data.

\label{sup:quantitative_performance_comparison_sigma_zero}

\begin{table}[h]
  \centering
  \vspace{-8pt}
  \tabcolsep=10pt
  \caption{Performance comparison on CUHK-PEDES(M), ICFG-PEDES(M),
RSTPReid(M), and UFineBench(M).}
  \vspace{-10pt}
    \resizebox{\linewidth}{!}{%
    \begin{tabular}{c|c|c|c|cc}
 \Xhline{1.5pt}
    \multirow{2}[1]{*}{Method} & CUHK  & ICFG  & RSTP  & \multicolumn{2}{c}{UFineBench} \\
          & Avg   & Avg   & Avg   & 6926-Avg. & 3C-Avg. \\
     \hline
    \multicolumn{1}{l|}{NAFS~\cite{NAFS}~\textcolor{gray}{(ECCV'18)}}  & - & - & - & 76.49  & 58.25  \\
    \multicolumn{1}{l|}{CMKA~\cite{CMKA}~\textcolor{gray}{(TIP'21)}}  & 70.07  & - & - & - & - \\
    \multicolumn{1}{l|}{LapsCore~\cite{LapsCore}~\textcolor{gray}{(ICCV'21)}} & 75.60  & - & - & - & - \\
    \multicolumn{1}{l|}{SSAN~\cite{SSAN}~\textcolor{gray}{(arXiv'21)}}  & - & - & - & 85.52  & 67.32  \\
     \multicolumn{1}{l|}{LGUR~\cite{LGUR}~\textcolor{gray}{(MM'22)}}  & - & - & - & 81.72  & 65.42  \\
    \multicolumn{1}{l|}{AXM-Net~\cite{AXM-Net}~\textcolor{gray}{(MM'22)}} & 77.24  & - & - & - & - \\
    \multicolumn{1}{l|}{SAF~\cite{SAF}~\textcolor{gray}{(ICASSP'22)}}   & 78.38  & - & - & - & - \\
    \multicolumn{1}{l|}{TIPCB~\cite{TIPCB}~\textcolor{gray}{(Neuro'22)}} & 78.85  & - & - & - & - \\
    \multicolumn{1}{l|}{MANet~\cite{MANet}~\textcolor{gray}{(TNNLS'23)}} & 79.14  & 73.00  & - & - & - \\
    \multicolumn{1}{l|}{CFine~\cite{CFine}~\textcolor{gray}{(TIP'23)}} & 82.22  & 73.27  & 68.22  & - & - \\
    \multicolumn{1}{l|}{IRRA~\cite{IRRA}~\textcolor{gray}{(CVPR'23)}}  & 85.67  & 76.51  & 76.57  & 90.81  & 68.99  \\
    \multicolumn{1}{l|}{BiLMa~\cite{BiLMa}~\textcolor{gray}{(ICCV'23)}} & 85.75  & 76.57  & 77.17  & - & - \\
    \multicolumn{1}{l|}{RaSa~\cite{RaSa}~\textcolor{gray}{(IJCAI'23)}}  & 87.02  & 76.93  & 81.58  & - & - \\
       \multicolumn{1}{l|}{IRRA~\cite{IRRA}~\textcolor{gray}{(CVPR'23)}} & 78.56  & 71.52  & 64.32  & 87.05  & - \\
    \multicolumn{1}{l|}{TBPS-CLIP~\cite{TBPS-CLIP}~\textcolor{gray}{(AAAI'24)}} & 84.69  & 76.95  & 78.08  & - & - \\
    \multicolumn{1}{l|}{CADA-G~\cite{CADA-G}~\textcolor{gray}{(TMM'24)}} & 85.72  & 75.71  & 77.75  & - & - \\
    \multicolumn{1}{l|}{UMSA~\cite{UMSA}~\textcolor{gray}{(AAAI'24)}}  & 85.89  & 77.33  & 79.00  & - & - \\
    \multicolumn{1}{l|}{FSRL~\cite{FSRL}~\textcolor{gray}{(ICMR'24)}}  & 86.32  & 77.28  & 77.77  & - & - \\
    \multicolumn{1}{l|}{Propot~\cite{Propot}~\textcolor{gray}{(MM'24)}} & 86.32  & 77.89  & 78.40  & - & - \\
    \multicolumn{1}{l|}{RDE~\cite{RDE}~\textcolor{gray}{(CVPR'24)}}   & 86.73  & 79.17  & 79.73  & - & - \\
    \multicolumn{1}{l|}{CFAM~\cite{CFAM}~\textcolor{gray}{(CVPR'24)}}  & 86.83  & 77.63  & 79.03  & 93.86  & \underline{74.63}  \\
    \multicolumn{1}{l|}{Bi-IRRA~\cite{Bi-IRRA}~\textcolor{gray}{(TPAMI'25)}} & \underline{88.77}  & \underline{79.79}  & \underline{84.17}  & \underline{95.18}  & - \\
    \hline
    \rowcolor[rgb]{0.910, 0.941, 0.980}  \multicolumn{1}{l|}{\textbf{LightAIR (Ours)}}  &   \textbf{89.12}    &   \textbf{80.15}    &   \textbf{85.55}    &   \textbf{95.37}    & \textbf{83.32} \\
    \Xhline{1.5pt}
    \end{tabular}%
    }
  \label{tab:TIPR}%
  \vspace{-15pt}
\end{table}%

\begin{figure*}[ht]
    \centering
	\includegraphics[width=0.9\linewidth]{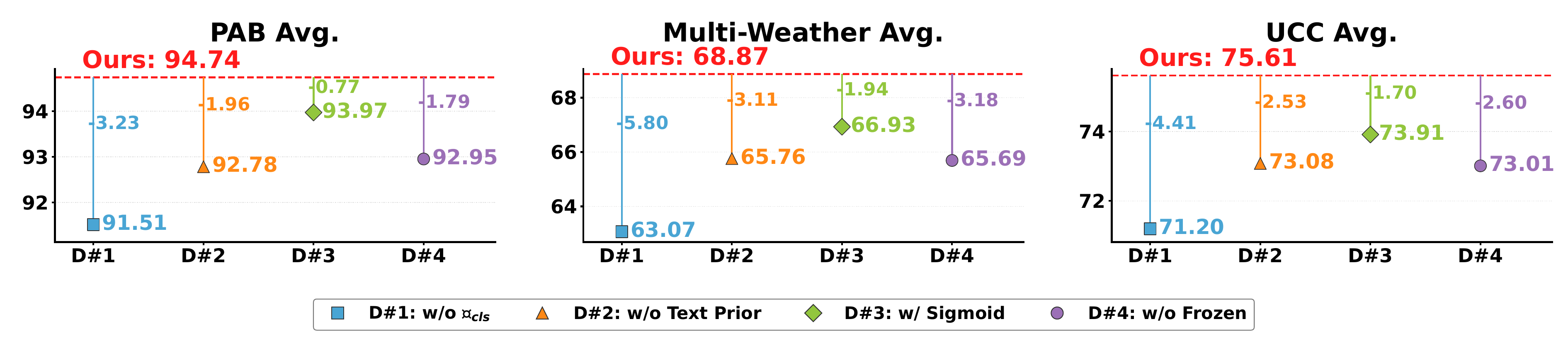}
    \vspace{-10pt}
	\caption{Ablation study of AIO module on PAB, Multi-Weather, and UCC datasets. Figure best viewed in color.}    
    \vspace{-6pt}
	\label{fig:AIO}
\end{figure*}

\subsubsection{On TIPR Task}

\noindent To further verify the generalization of the LightAIR framework, we additionally evaluate its performance on the conventional Text-to-Image Person Retrieval (TIPR) task. As shown in Table~\ref{tab:TIPR}, LightAIR achieves the best performance across four mainstream TIPR benchmarks (CUHK-PEDES, ICFG-PEDES, RSTPReid, and UFineBench). Notably, in the most challenging UFine3C evaluation containing abundant complex visual features, the average Recall of LightAIR achieves a significant improvement of 8.69\%. This fully demonstrates the excellent generalization of LightAIR. In TIPR, features extracted by the AIO module serve as fine grained supplementary information beyond identity in traditional TIPR, achieving fine-grained feature retrieval. Simultaneously, the GR module universally generalizes to conventional TIPR scenarios, ensuring the stability of the model learning process.

\subsection{Ablation Study}
We evaluate each component on PAB, Multi-Weather, and UCC; efficiency and additional results are provided in Appendix~\ref{sup:efficiency} and Appendix~\ref{sup:addtional_ablation_study}.

\subsubsection{Action Inversion Operator (AIO)}

Figure~\ref{fig:AIO} verifies the AIO design. Removing $\mathcal{L}_{cls}$ produces the largest Multi-Weather decline (5.80 points), while removing the text prior or codebook freezing, or replacing Softmax with Sigmoid, consistently reduces performance. These results support explicit semantic supervision and a stable, sparse semantic anchor.

\subsubsection{Orthogonal Null-Space Projection (ONSP)}

\begin{figure}[h]
    \centering
    \vspace{-8pt}
	\includegraphics[width=\linewidth]{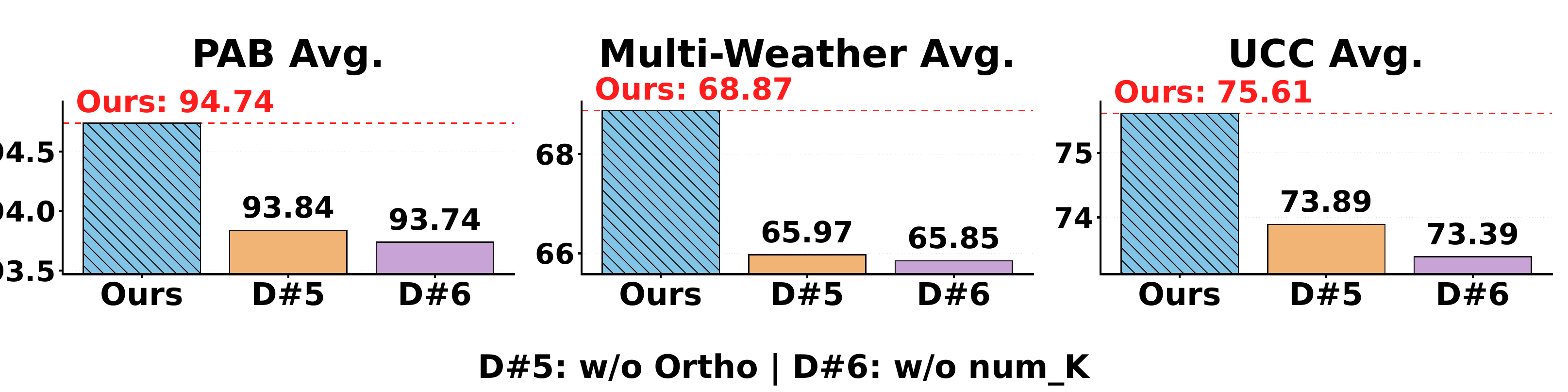}
    \vspace{-20pt}
	\caption{Ablation study of ONSP module on PAB, Multi-Weather, and UCC datasets. Figure best viewed in color.}
    \vspace{-14pt}
	\label{fig:ONSP}
\end{figure}

Figure~\ref{fig:ONSP} shows that replacing null-space projection with feature subtraction or using multiple action bases lowers Multi-Weather R@1 by 2.90 and 3.02 points, respectively. This supports strict orthogonal decoupling with a focused action basis.

\subsubsection{Gradient Rectification (GR)}

\begin{figure}[h]
    \centering
    \vspace{-6pt}
	\includegraphics[width=\linewidth]{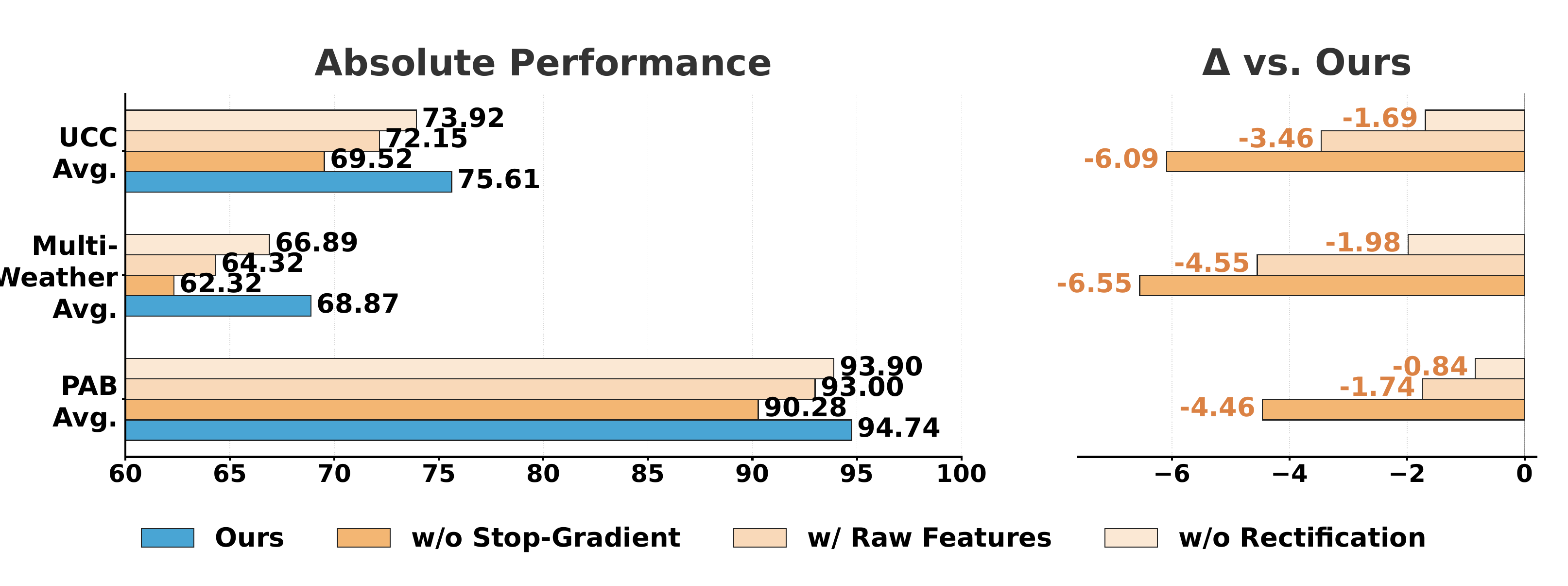}
      \vspace{-20pt}
	\caption{Ablation study of GR module on PAB, Multi-Weather, and UCC datasets. Figure best viewed in color.}
      \vspace{-15pt}
	\label{fig:GR}
\end{figure}

Figure~\ref{fig:GR} confirms the importance of controlling the optimization path. Removing Stop-Gradient causes the largest Multi-Weather and UCC declines (6.55 and 6.09 points), while raw features and unrectified gradients also degrade retrieval. This indicates that gradient control prevents harmful shortcuts during alignment.

\subsubsection{Entropy-Guided Retraction}

\begin{figure}[h]
    \centering

\includegraphics[width=0.9\linewidth]{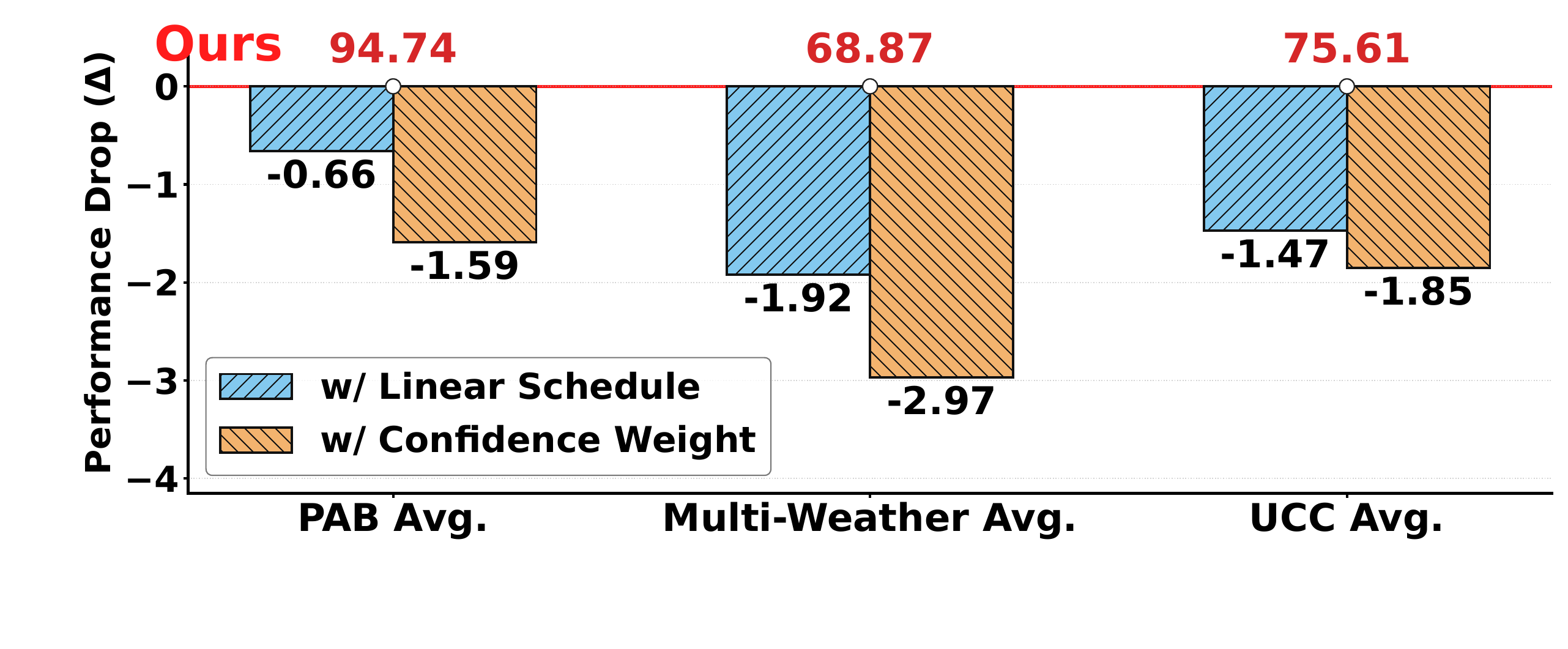}
      \vspace{-10pt}
	\caption{Ablation study of Entropy-Guided Retraction process. Figure best viewed in color.}
        \vspace{-10pt}	
	\label{fig:Omega}
\end{figure}

Figure~\ref{fig:Omega} shows that confidence-only weighting and a linear schedule reduce Multi-Weather performance by 2.97 and 1.92 points, respectively, confirming the value of entropy-guided adaptive retraction.

\subsection{Sensitivity Analysis}

\begin{figure}[h]
    \centering
    \vspace{-10pt}	\includegraphics[width=0.97\linewidth]{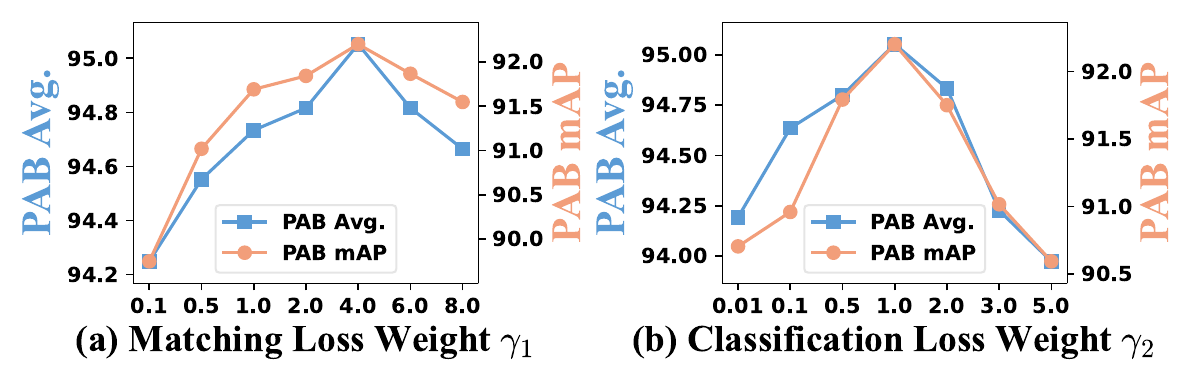}
      \vspace{-13pt}
	\caption{Hyper-parameters sensitivity analysis of loss weights on PAB dataset.}
  \vspace{-15pt}
	\label{fig:sensitivity}
\end{figure}

To analyze the sensitivity of the model to the image text matching loss weight $\gamma_1$ and the action discriminative loss weight $\gamma_2$, Figure~\ref{fig:sensitivity} displays the performance variation curves under different parameters. It can be observed that as both parameters increase, model performance exhibits a trend of initially rising and then falling. Regarding the image text matching loss $\mathcal{L}_{itm}$, model performance improves with the increase of $\gamma_1$ and achieves the optimum at $\gamma_1=4.0$. This indicates that moderately enhancing fine-grained cross-modal matching supervision helps improve the accuracy of visual feature geometric decoupling. However, when $\gamma_1 > 4.0$, performance begins to decline, indicating that excessive matching constraints lead to feature space over-regularization, thereby restricting the visual diversity required for fine grained retrieval. For the action discriminative loss $\mathcal{L}_{cls}$, $\gamma_2=1.0$ achieves the optimal balance on both datasets. Specifically, a smaller $\gamma_2$ (e.g., $0.1$) makes it difficult for the action mapping network to effectively extract weak action signals, causing the model to remain constrained by excessively strong appearance features in the visual space. Conversely, a larger $\gamma_2$ (e.g., $5.0$) will dominate the optimization process, causing the model to deviate from the core retrieval objective. 

\subsection{Case Study}

\begin{figure}[h]
    \centering
	\includegraphics[width=0.97\linewidth]{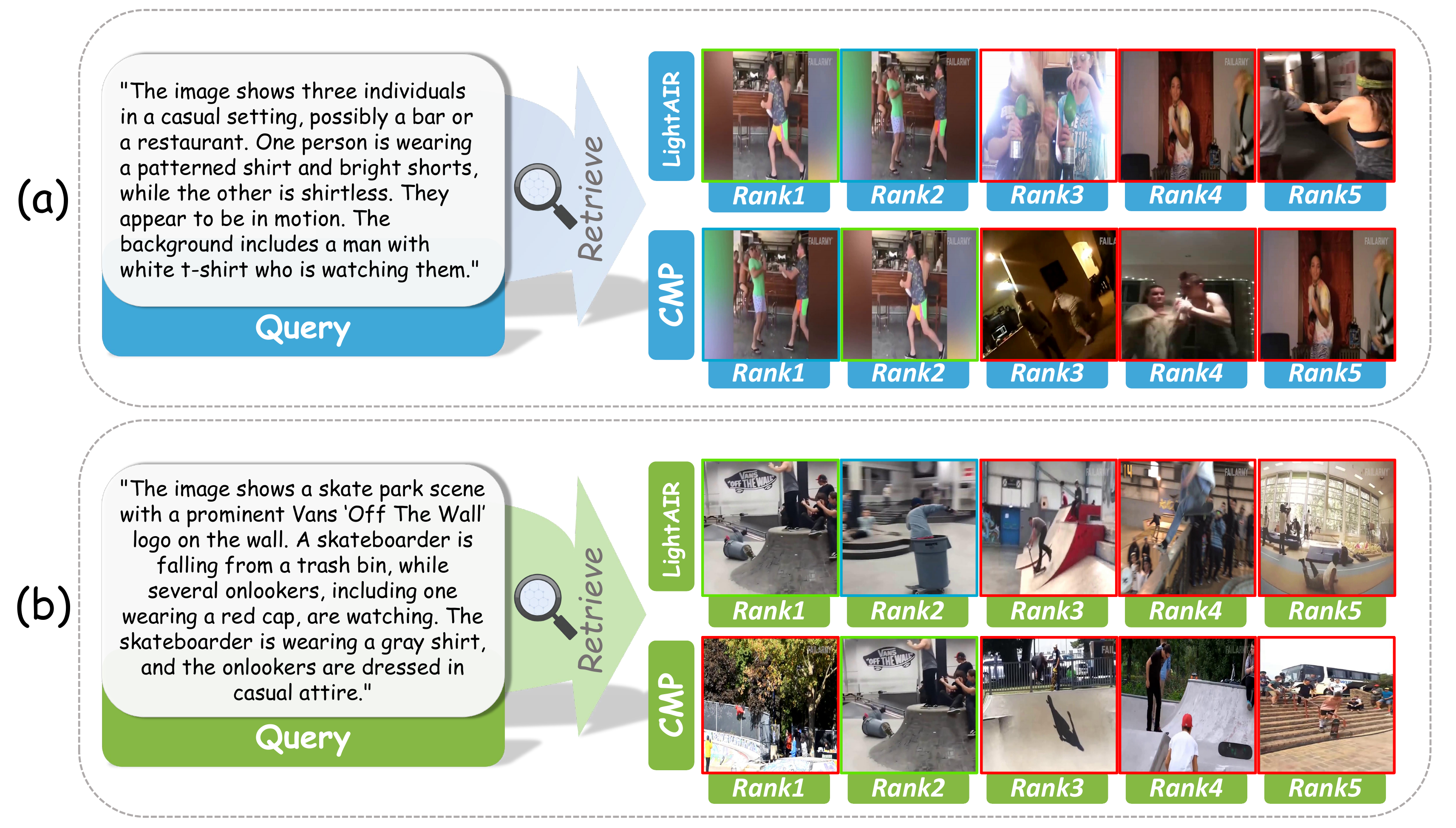}
      \vspace{-5pt}
	\caption{Case study on LightAIR and CMP. Matched images are marked by green boxes, mismatched images are marked in red, and blue boxes indicate the hard negatives.}
  \vspace{-3pt}
	\label{fig:Case}
\end{figure}

To qualitatively evaluate the retrieval performance of LightAIR, Figure~\ref{fig:Case} presents a visual comparison with the representative baseline CMP on the PAB dataset. We obtain the following observations:
1) Figure~\ref{fig:Case}(a) examines the fine-grained alignment capability of the model in highly similar scenes. Given a query containing texts such as ``patterned shirt'', ``shirtless'', and ``in motion'', the Top-1 result of CMP fits the macro-level scene distribution but suffers from misalignment between local appearance and action, exposing the defect that its implicit action representation is highly susceptible to being engulfed by appearance information. In contrast, LightAIR retrieves correct target image, verifying the effectiveness of AIO module in robustly extracting action signals utilizing text semantic anchors, while the ONSP module achieves orthogonal decoupling of both, preventing feature contamination.
2) Figure~\ref{fig:Case}(b) highlights the robustness against hard negative samples. Faced with fine grained queries like ``falling from a trash bin'', although CMP captures the main action, it ignores local entity constraints and erroneously recalls interfering samples with identical actions but entirely different appearance features (e.g., ``red jacket''). The accurate hit of LightAIR strongly proves that the dynamic Riemannian gradient rectification of the GR module successfully blocks the shortcut learning of forcibly attributing appearance differences to action differences, and fundamentally suppressing the misleading of interfering samples in complex scenes.

%% file: 5_con.tex
\section{Conclusion}
In this paper, we proposed LightAIR to tackle the critical challenges of visual decoupling failure and shortcut learning in Text-based Person Anomaly Search (TPAS). By integrating an \textit{Action Inversion Operator} with \textit{Orthogonal Null-Space Projection}, our framework achieved mathematically guaranteed action-appearance decoupling without relying on fragile external pose estimators. Additionally, a novel \textit{Gradient Rectification} module constrained backpropagation along the Riemannian tangent space to prevent manifold drift during hard negative optimization. Extensive experiments on the TPAS benchmark and traditional TIPR benchmarks demonstrated that LightAIR achieves robust cross-modal alignment and significantly outperforms existing state-of-the-art methods.

%% file: X_suppl.tex
\clearpage

\onecolumn 
\setcounter{page}{1} 
\renewcommand{\thepage}{S\arabic{page}} 


\twocolumn[{ 
  \centering
  \vspace*{1cm}
  
  {\Huge \bfseries Supplementary Material \par} 
  
  \vspace{0.5cm}
  
  {\Large for \textit{\textbf{``LightAIR: Lightweight Action Inversion and Riemannian Rectification \\for Text-based Person Anomaly Search''}} \par}
  
  \vspace{1cm}
}]

\noindent
\appendix
\setcounter{page}{1}
\setcounter{equation}{0}

\noindent This is the appendix of ``LightAIR: Lightweight Action Inversion and Riemannian Rectification for Text-based Person Anomaly Search''. 
\begin{itemize}
    \item \textbf{Appendix~\ref{sup:datasets}}: Datasets
    \begin{itemize}
        \item \textbf{Appendix~\ref{sup:tpas}}: Datasets For TPAS
        \item \textbf{Appendix~\ref{sup:tipr}}: Datasets For TIPR
    \end{itemize}
    \item \textbf{Appendix~\ref{sup:addtional_quantitative_analysis}}: Additional Quantitative Analysis 
    \begin{itemize}
        \item \textbf{Appendix~\ref{sup:complete_tipr}}: Complete TIPR
        \item \textbf{Appendix~\ref{sup:efficiency}}: Efficiency Evaluation
        \item \textbf{Appendix~\ref{sup:sensi_K}}: Sensitivity Analysis of $K$
    \end{itemize}
    \item \textbf{Appendix~\ref{sup:addtional_ablation_study}}: Additional Ablation Study
    \begin{itemize}
        \item \textbf{Appendix~\ref{sup:complete_ablation_study}}: Complete Ablation Study
    \end{itemize}
    \item \textbf{Appendix~\ref{sup:training_procedure}}: Algorithm of Training Procedure
    \item \textbf{Appendix~\ref{sup:more_case}}: More Case Study
\end{itemize}

\section{Datasets}
\label{sup:datasets}
To fully evaluate the performance of the proposed LightAIR model, we conducted extensive experiments on multiple mainstream benchmarks covering two major tasks: Text-based Person Anomaly Search (TPAS) and traditional Text-to-Image Person Retrieval (TIPR). Specific details of each dataset and evaluation setting are described below.

\subsection{Datasets For TPAS.}
\label{sup:tpas}
For the TPAS task, our evaluation mainly relies on the large scale Pedestrian Anomaly Behavior (PAB) benchmark dataset, combined with two extended settings, Multi-weather and UCC, to examine model robustness under complex environments and unknown distributions.

\textbf{PAB dataset.}
This benchmark is primarily used to fill the gap of lacking abnormal behavior descriptions in existing retrieval datasets. Both its training and test sets are constructed based on the OOPS! video library. During the training phase, we utilized synthetic data containing 1,013,605 image text pairs, involving 1,000 categories of normal actions (e.g., running, playing football) and 1,600 categories of abnormal events (e.g., falling, being hit). Its test set consists of 1,978 real scene images (namely 989 matched pairs), strictly maintaining a balanced 1:1 ratio of normal to abnormal behavior samples.

\textbf{Multi-weather Setting.}
To test model stability under extreme weather conditions, this is a test benchmark specifically designed for robustness evaluation. Based on the PAB real world test set, this setting injects 10 categories of environmental interference simulations, including wind, rain, snow, dark, dark with wind, dark with rain, dark with snow, and overexposure.

\textbf{OOD Setting.}
This test set is specifically used to evaluate the out of distribution (OOD) generalization level of the model, for which we adopted the UCC dataset. Its data is independently collected from the UCF-Crime video library, and its data distribution differs from PAB. By extracting keyframes from videos across 13 different categories of abnormal and normal behaviors and utilizing Qwen2-VL to generate corresponding text descriptions, 5,320 independent image text pairs were finally compiled.

\subsection{Datasets For TIPR}
\label{sup:tipr}
For the traditional TIPR task, we selected four widely adopted benchmark datasets: CUHK-PEDES, ICFG-PEDES, RSTPReid, and UFineBench, to verify the general retrieval capability of the model.

\textbf{CUHK-PEDES.}
As a classic benchmark in this field, it contains 40,206 images and 80,440 text descriptions of 13,003 identities (IDs). Following the conventional split protocol, 34,054 images containing 11,003 IDs are used for training, and 3,074 images containing 1,000 IDs are used for testing.

\textbf{ICFG-PEDES.}
This dataset includes 54,522 images of 4,102 IDs, with each image corresponding to a single text annotation. According to the standard protocol, the training set is allocated 34,674 images (3,102 IDs), and the test set is allocated 19,848 images (1,000 IDs).

\textbf{RSTPReid.}
Images in this dataset are captured by 15 independent cameras, covering 20,505 images of 4,101 IDs. Each ID contains exactly 5 images, and each image is accompanied by 2 text descriptions. The dataset is divided into 3,701 IDs for training, 200 IDs for validation, and 200 IDs for testing.

\textbf{UFineBench.}
This is the latest benchmark focusing on ultra fine grained text to image person retrieval. Its core subset UFine6926 contains 26,206 images and 52,412 detailed descriptions of 6,926 IDs, with an average text length of 80.8 words, far exceeding the granularity of existing datasets. The data is divided into a training set of 18,577 images (4,926 IDs) and a test set of 7,629 images (2,000 IDs). Furthermore, we also tested the model utilizing the special UFine3C evaluation set. This subset integrates test data from multiple aforementioned datasets and expands query diversity via large language models (LLMs) to evaluate cross domain, cross granularity, and cross style retrieval performance.

\section{Additional Quantitative Analysis}
\label{sup:addtional_quantitative_analysis}

\begin{table*}[htbp]
  \centering
  \caption{{Complete Performance Comparisons on CUHK-PEDES(M), ICFG-PEDES(M) and RSTPReid(M)}}
  \vspace{-10pt}
  \resizebox{\linewidth}{!}{%
    \begin{tabular}{l|cccc|cccc|cccc}
    \hline
    \hline
    \multicolumn{1}{c|}{\multirow{2}{*}{Methods}} & \multicolumn{4}{c|}{CUHK-PEDES} & \multicolumn{4}{c|}{ICFG-PEDES} & \multicolumn{4}{c}{RSTPReid} \\
          & \multicolumn{1}{c}{R@1 } & \multicolumn{1}{c}{R@5} & \multicolumn{1}{c}{R@10} & \multicolumn{1}{c|}{R-Avg} & \multicolumn{1}{c}{R@1 } & \multicolumn{1}{c}{R@5} & \multicolumn{1}{c}{R@10} & \multicolumn{1}{c|}{R-Avg} & \multicolumn{1}{c}{R@1 } & \multicolumn{1}{c}{R@5} & \multicolumn{1}{c}{R@10} & \multicolumn{1}{c}{R-Avg} \\
    \hline
    CMKA~\textcolor{gray}{(TIP'21)} & 54.69  & 73.65  & 81.86  & 70.07  & -     & -     & -     & -     & -     & -     & -     & - \\
    LapsCore~\textcolor{gray}{(ICCV'21)} & 63.40  & -     & 87.80  & 75.60  & -     & -     & -     & -     & -     & -     & -     & - \\
    SAF~\textcolor{gray}{(ICASSP'22)} & 64.13  & 82.62  & 88.40  &78.38     & -     & -     & -     & -     & -     & -     & -     & - \\
    TIPCB~\textcolor{gray}{(Neuro'22)} & 64.26  & 83.19  & 89.10  &78.85   & -     & -     & -     & -     & -     & -     & -     & - \\
    AXM-Net~\textcolor{gray}{(MM'22)} & 64.44  & 80.52  & 86.77  & 77.24   & -     & -     & -     & -     & -     & -     & -     & - \\
    MANet~\textcolor{gray}{(TNNLS'23)} & 65.64  & 83.01  & 88.78  & 79.14    & 59.44  & 76.80  & 82.75  & 73.00   & -     & -     & -     & - \\
    CFine~\textcolor{gray}{(TIP'23)} & 69.57  & 85.93  & 91.15  & 82.22     & 60.83  & 76.55  & 82.42  & 73.27    & 50.55  & 72.50  & 81.60  &68.22 \\
    IRRA~\textcolor{gray}{(CVPR'23)} & 73.38  & 89.93  & 93.71  & 85.67   & 63.46  & 80.25  & 85.82  & 76.51   & 60.20  & 81.30  & 88.20  & 76.57   \\
    BiLMa~\textcolor{gray}{(ICCV'23)} & 74.03  & 89.59  & 93.62  & 85.75   & 63.83  & 80.15  & 85.74  & 76.57  & 61.20  & 81.50  & 88.80  &77.17  \\
    RaSa~\textcolor{gray}{(IJCAI'23)} & 76.51  & 90.29  & 94.25  & 87.02 & 65.28  & 80.40  & 85.12  & 76.93   & 66.90  & 86.50  & 91.35  & 81.58   \\
    TBPS-CLIP~\textcolor{gray}{(AAAI'24)} & 73.54  & 88.19  & 92.35  & 84.69  & 65.05  & 80.34  & 85.47  & 76.95  & 61.95  & 83.55  & 88.75  & 78.08   \\
    CADA-G~\textcolor{gray}{(TMM'24)} & 73.48  & 89.57  & 94.10  &85.72 
  & 62.54  & 79.46  & 85.14  & 75.71   & 61.50  & 82.60  & 89.15  & 77.75 \\
    UMSA~\textcolor{gray}{(AAAI'24)} & 74.25  & 89.83  & 93.58  & 85.89 
  & 65.62  & 80.54  & 85.83  & 77.33  & 63.40  & 83.30  & 90.30  &79.00   \\
    FSRL~\textcolor{gray}{(ICMR'24)} & 74.86  & 89.97  & 94.14  & 86.32 
  & 64.93  & 80.71  & 86.19  & 77.28  & 60.65  & 83.05  & 89.60  &77.77   \\
    Propot~\textcolor{gray}{(MM'24)} & 74.89  & 89.90  & 94.17  & 86.32 
  & 65.12  & 81.57  & 86.97  &77.89   & 61.87  & 83.63  & 89.70  & 78.40   \\
    CFAM~\textcolor{gray}{(CVPR'24)} & 75.60  & 90.53  & 94.36  & 86.83 
  & 65.38  & 81.17  & 86.35  & 77.63  & 62.45  & 83.55  & 91.10  & 79.03  \\
    RDE~\textcolor{gray}{(CVPR'24)} & 75.94  & 90.14  & 94.12  & 86.73 
  & 67.68  & 82.47  & 87.36  & 79.17 & 65.35  & 83.95  & 89.90  &79.73  \\
    Bi-IRRA~\textcolor{gray}{(TPAMI'25)} & 78.82  & 92.02  & 95.47  & 88.77 
  & 68.53  & 83.04  & 87.79  & 79.79   & 72.85  & 87.75  & 91.90  & 84.17  \\
    \hline
    \rowcolor[rgb]{0.910, 0.941, 0.980}
    \textbf{LightAIR (Ours)}  &  \textbf{78.91} &    \textbf{92.57} &   \textbf{95.89} &  	\textbf{89.12} & \textbf{69.43} &	\textbf{83.23} &	\textbf{87.79} &	\textbf{80.15} &  \textbf{73.95} &	\textbf{88.95} & \textbf{93.75} &	\textbf{85.55} \\
    \hline
    \hline
    \end{tabular}%
    }
  \vspace{-10pt}
  \label{tab:complete_tipr}%
\end{table*}%

\subsection{Complete TIPR}
\label{sup:complete_tipr}

To comprehensively evaluate the performance of LightAIR on the traditional Text-to-Image Person Retrieval (TIPR) task, we conducted exhaustive experiments on four mainstream benchmarks: CUHK-PEDES, ICFG-PEDES, RSTPReid. As shown in Table~\ref{tab:complete_tipr}, LightAIR achieves SOTA performance across all evaluation metrics (R@1, R@5, R@10, R-Avg) on the three datasets, outperforming existing baseline methods including the second best models Bi-IRRA~\cite{Bi-IRRA} and CFAM~\cite{CFAM}.

Experimental results show that LightAIR exhibits stable retrieval superiority across all evaluation benchmarks. On CUHK-PEDES, LightAIR reaches an R@1 of 78.91\%, achieving a +0.09\% improvement over the previous Bi-IRRA. On ICFG-PEDES and RSTPReid, LightAIR obtains R@1 improvements of 69.43\% (+0.90\% over Bi-IRRA) and 73.95\% (+1.10\% over Bi-IRRA), respectively.

The performance improvement of LightAIR stems from its effective response to the two core challenges in the TPAS task. Due to the lack of explicit mathematical constraints, existing soft decoupling methods struggle to separate pixel level highly coupled appearance and actions, causing weak action features to be easily contaminated. Moreover, when facing hard negative samples with identical appearances but different behaviors, the conventional optimization process is susceptible to excessive gradient penalties and falls into optimization shortcuts, which forcibly create semantic differences by distorting underlying mapping rules, thereby causing action semantic manifold drift.

LightAIR achieves more precise retrieval through the close synergy of three major modules. In the forward representation phase, the model first combines the text semantic prior codebook and latent action coefficient estimation to reconstruct a reliable action representation $\mathbf{z}_{act}$ from the highly entangled visual space, compensating for the unreliability of pure visual extraction. Subsequently, it strictly projects the global feature onto the null space of this action feature, physically peeling off the action component at the geometric level to obtain a high purity appearance feature $\mathbf{z}_{app}$. In the backpropagation phase, the model further orthogonally projects the Euclidean gradient onto the tangent space of the action semantic manifold, supplemented by Shannon entropy based adaptive weighting. This mechanism ensures that features are always updated along legitimate directions, effectively avoiding harmful optimization shortcuts triggered by hard negative samples from the root.

\begin{table*}[htbp]
  \centering
  \caption{Efficiency and Performance Comparison. PAB-Avg and UCC(OOD)-Avg are the mean of R@\{1,5,10\} in the 1M setting; MultiWeather reports mean R@1.}
  \vspace{-10pt}
    \resizebox{\linewidth}{!}{%
    \begin{tabular}{c|c|c|c|c|c|c|c}
    \Xhline{1.5pt}
    Method & Parameters(M) & GPU Memory(MiB) & Test time(s/sample) & Train Time(s/iteration) & PAB-Avg & MultiWeather R@1 & UCC(OOD)-Avg \\
    \hline
    CMP  & 226.13 & 14575(bs=22) & 0.0015 & 0.5326(bs=22) & 94.59 & 67.12 & 68.30  \\
    LightAIR (Ours) & 233.06 & 19543(bs=22) & 0.0023 & 0.4342(bs=22) & 95.04 & 68.87 & 76.22  \\
    \Xhline{1.5pt}
    \end{tabular}%
    }
  \label{tab:quantitative_efficiency}%
\end{table*}%

\subsection{Efficiency Evaluation}
\label{sup:efficiency}

To evaluate system efficiency and resource consumption, we conduct efficiency evaluation experiments. Table~\ref{tab:quantitative_efficiency} details the multidimensional comparison metrics between LightAIR and the baseline model CMP regarding model complexity, GPU memory footprint, and runtime efficiency. The specific analysis is as follows:

Regarding model complexity, although rigorous mathematical projection and decoupling operations are introduced, benefiting from the lightweight network design, the total parameter count of LightAIR (233.06M) exhibits only a marginal increase over the baseline (226.13M), effectively controlling the overall model complexity.
Regarding runtime efficiency, despite the GPU memory footprint increasing to 19543 MiB and the test time slightly rising to 0.0023s/sample under the same batch size (bs=22), it is noteworthy that the single step training time of LightAIR is merely 0.4342s/iter, significantly faster than 0.5326s/iter of the baseline. This phenomenon is attributed to the AIO and ONSP modules utilizing explicit orthogonal geometric projections to replace the inefficient parameter fitting process in traditional black box paradigms. Simultaneously, the GR module effectively suppresses harmful optimization shortcuts during backpropagation, avoiding invalid parameter updates and feature semantic distortions, thereby accelerating the model convergence process and training efficiency.

Furthermore, the modest computational overhead yields clear performance gains. LightAIR maintains the lead on PAB-Avg (95.04\% vs 94.59\%) and improves MultiWeather R@1 (68.87\% vs 67.12\%) and UCC(OOD)-Avg (76.22\% vs 68.30\%), demonstrating stronger robustness under background noise and distribution shifts.

\subsection{Sensitivity Analysis of $K$}
\label{sup:sensi_K}

\begin{figure}[ht]
    \centering
	\includegraphics[width=0.9\linewidth]{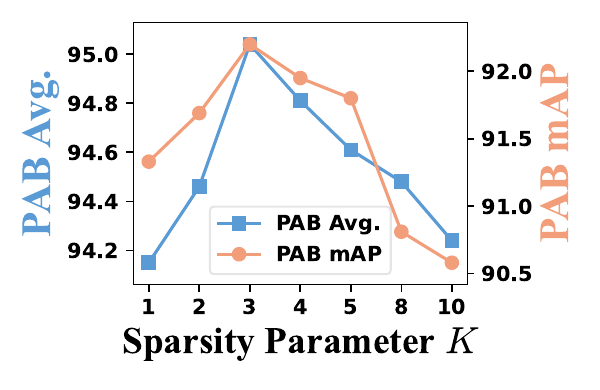}
    \vspace{-10pt}
	\caption{Sensitivity analysis of the sparsity parameter $K$ in the AIO module.}
	\label{fig:sensitivity_k}
    \vspace{-10pt}
\end{figure}

To analyze the sensitivity of the model to the retained parameter $K$ in the Top-K sparsification operation, Figure~\ref{fig:sensitivity_k} presents the corresponding performance variation curve. It can be observed that as the value of $K$ increases, the model performance exhibits a trend of initially rising and then falling, achieving the optimum at $K=3$. This phenomenon highly aligns with the design mechanism of the Latent Coefficient Estimator. Specifically, when $K$ is small (e.g., $1$ or $2$), the constraint of the network as an information isolation bottleneck is overly strict, and the coordinate coefficients mapped into the action semantic subspace are insufficient, making the model unable to accurately describe complex actions in reality. Conversely, when $K > 3$, the performance begins to gradually decline, indicating that an excessively large parameter weakens the sparsification effect of the Top-K mechanism. It not only fails to suppress the interference of irrelevant redundant action semantics but also easily reintroduces previously isolated visual appearance noise into the reconstruction process. Therefore, $K=3$ appropriately prompts the model to utilize only a few core action components to accurately describe the current complex action, achieving an optimal balance between semantic expression capacity and noise suppression.

\section{Additional Ablation Study}
\label{sup:addtional_ablation_study}

\subsection{Complete Ablation Study}
\label{sup:complete_ablation_study}

\begin{table}[htbp]
  \centering
  \caption{Ablation study for LightAIR on the PAB dataset (1M training setting).}
  \vspace{-10pt}
  \label{tab:ablation_lightair_pab}
  \resizebox{0.9\linewidth}{!}{
  \begin{tabular}{l|l|cccc}
    \Xhline{1.5pt} 
    \textbf{D\#} & \textbf{Derivatives} & \textbf{R@1} & \textbf{R@5} & \textbf{R@10} & \textbf{mAP} \\
    \Xhline{0.8pt} 
    (1) & w/o $L_{cls}$ & 81.45 & 96.39 & 96.70 & 87.87 \\
    (2) & w/o Text Prior & 82.71 & 97.49 & 98.15 & 89.89 \\
    (3) & w/ Sigmoid & 83.56 & 98.65 & 99.70 & 90.69 \\
    (4) & w/o Frozen & 82.00 & 97.60 & 99.25 & 89.57 \\
    (5) & w/o Ortho & 83.24 & 98.75 & 99.55 & 90.38 \\
    (6) & w/o $num_K$ & 83.08 & 98.60 & 99.55 & 89.60 \\
    (7) & w/o Stop-Gradient & 80.55 & 95.44 & 94.85 & 84.91 \\
    (8) & w/ Raw Feature & 82.22 & 97.60 & 99.20 & 89.34 \\   
    (9) & w/o Rectification & 83.42 & 98.65 & 99.65 & 90.29 \\
    (10) & w/ Linear Schedule & 83.92 & 99.02 & 99.30 & 90.81 \\
    (11) & w/ Confidence Weight & 81.83 & 98.70 & 98.92 & 90.23 \\
    \Xhline{0.8pt} 
    \rowcolor[rgb]{0.910, 0.941, 0.980}
    \multicolumn{2}{c|}{\textbf{LightAIR (Ours)}} & \textbf{85.49} & \textbf{99.65} & \textbf{99.99} & \textbf{92.20} \\
    \Xhline{1.5pt} 
  \end{tabular}
  }
  \vspace{-10pt}
\end{table}

\begin{table}[htbp]
  \centering
  \caption{Ablation study for LightAIR under the corrected Multi-weather setting.}
  \vspace{-10pt}
  \label{tab:ablation_lightair_full}
  \resizebox{\linewidth}{!}{
  \begin{tabular}{l|l|cc}
    \Xhline{1.5pt}
    \textbf{D\#} & \textbf{Derivatives} & \textbf{Mean R@1} & \textbf{Mean mAP} \\
    \hline\hline
    (1) & w/o $\mathcal{L}_{cls}$ & 63.07 & 74.97 \\
    (2) & w/o Text Prior & 65.76 & 77.81 \\
    (3) & w/ Sigmoid & 66.93 & 78.92 \\
    (4) & w/o Frozen & 65.69 & 77.74 \\
    (5) & w/o Ortho & 65.97 & 78.04 \\
    (6) & w/o $num_K$ & 65.85 & 77.79 \\
    (7) & w/o Stop-Gradient & 62.32 & 73.30 \\
    (8) & w/ Raw Feature & 64.32 & 76.43 \\
    (9) & w/o Rectification & 66.89 & 78.92 \\
    (10) & w/ Linear Schedule & 66.95 & 77.88 \\
    (11) & w/ Confidence Weight & 65.90 & 79.09 \\
    \hline
    \rowcolor[rgb]{0.910, 0.941, 0.980}
    \multicolumn{2}{c|}{\textbf{LightAIR (Ours)}} & \textbf{68.87} & \textbf{79.69} \\
    \Xhline{1.5pt}
  \end{tabular}
  }
  \vspace{-10pt}
\end{table}

\begin{table}[htbp]
  \centering
  \caption{Ablation study for LightAIR under OOD setting on the UCC dataset (1M training setting).}
  \label{tab:ablation_lightair_ucc}
  \vspace{-10pt}
  \resizebox{\linewidth}{!}{
  \begin{tabular}{l|l|cccc}
    \Xhline{1.5pt}
    \textbf{D\#} & \textbf{Derivatives} & \multicolumn{1}{c}{\textbf{R@1}} & \multicolumn{1}{c}{\textbf{R@5}} & \multicolumn{1}{c}{\textbf{R@10}} & \multicolumn{1}{c}{\textbf{mAP}} \\
    \hline\hline
    (1) & w/o $\mathcal{L}_{cls}$ & 56.13 & 74.96 & 82.50 & 48.63 \\
    (2) & w/o Text Prior & 59.35 & 75.46 & 84.43 & 49.71 \\
    (3) & w/ Sigmoid & 60.34 & 76.27 & 85.13 & 50.53 \\
    (4) & w/o Frozen & 59.47 & 75.14 & 84.43 & 49.19 \\
    (5) & w/o Ortho & 60.00 & 76.32 & 85.35 & 50.15 \\
    (6) & w/o $num_K$ & 59.41 & 75.50 & 85.26 & 49.41 \\
    (7) & w/o Stop-Gradient & 53.48 & 73.31 & 81.78 & 47.39 \\
    (8) & w/ Raw Feature & 56.32 & 74.64 & 85.50 & 49.41 \\
    (9) & w/o Rectification & 60.47 & 76.05 & 85.23 & 49.45 \\
    (10) & w/ Linear Schedule & 60.96 & 76.11 & 85.35 & 50.68 \\
    (11) & w/ Confidence Weight & 60.14 & 76.02 & 85.12 & 49.81 \\
    \hline
    \rowcolor[rgb]{0.910, 0.941, 0.980}
    \multicolumn{2}{c|}{\textbf{LightAIR (Ours)}} & \textbf{63.27} & \textbf{78.63} & \textbf{86.76} & \textbf{52.25} \\
    \Xhline{1.5pt}
  \end{tabular}
  }
  \vspace{-10pt}
\end{table}

To evaluate the component effects of LightAIR, we conduct detailed 1M-setting ablation studies on PAB, Multi-Weather, and UCC. The visual 0.1M comparisons used in the main text are reported separately in the figure captions.

\textbf{G[A]: \textit{Action Inversion Operator (AIO)}.}
\textbf{D\#(1) w/o $\mathcal{L}_{cls}$}: Remove the action discriminative loss; \textbf{D\#(2) Text Prior}: Discard the semantic codebook distilled from text in favor of a randomly initialized parameter matrix as the codebook; \textbf{D\#(3) w/ Sigmoid}: Replace Softmax for calculating the activation coefficient with Sigmoid; \textbf{D\#(4) w/o Frozen}: Unfreeze the codebook encoder to allow it to dynamically update with the network.
\textbf{G[B]: \textit{Orthogonal Null-Space Projection (ONSP)}.}
\textbf{D\#(5) w/o Ortho}: Replace null-space projection with simple feature subtraction; \textbf{D\#(6) w/o $num_K$}: Omit the Top-K sparsification operation.
\textbf{G[C]: \textit{Gradient Rectification (GR)}.}
\textbf{D\#(7) w/o Stop-Gradient}: Remove the gradient clipping restriction preventing the appearance gradient from overstepping and tampering with the action operator; \textbf{D\#(8) w/ Raw Feature}: Omit decoupled feature recombination and directly use the raw visual feature; \textbf{D\#(9) w/o Rectification}: Completely remove the Riemannian gradient to degrade to updating solely with the Euclidean gradient.
\textbf{G[D]: \textit{Entropy-Guided Retraction}.} \textbf{D\#(10) w/ Linear Schedule}: Linearly increase $\omega$ with epochs; \textbf{D\#(11) w/ Confidence Weight}: Directly use the maximum probability as the weight instead of Shannon entropy.

\textbf{Component effectiveness on the PAB benchmark.}
The detailed data in Table~\ref{tab:ablation_lightair_pab} reveals the effectiveness of each component in LightAIR:
1) \textbf{Semantic estimation alignment precisely reconstructs pure action features.} Removing the action discriminative loss (D\#1) causes R@1 to drop to 81.45\%. This indicates that in a highly coupled visual space, explicit cross-modal semantic guidance plays a crucial role in accurately reconstructing action features.
2) \textbf{Static priors ensure the stability of semantic anchors.} Replacing the semantic codebook with a randomly initialized parameter matrix (D\#2) or unfreezing the codebook encoder (D\#4) causes R@1 to decrease to 82.71\% and 82.00\%, respectively. This demonstrates that random initialization or dynamic updating of the semantic codebook will destroy the representation purity of semantic anchors, thereby introducing irrelevant noise or visual interference.
3) \textbf{Orthogonal geometric decoupling significantly outperforms linear subtraction.} If the orthogonal geometric decoupling in the ONSP module degrades into simple linear feature subtraction (D\#5), the model will trigger feature contamination due to the lack of strict geometric constraints, leading to an R@1 drop to 83.24\%. This proves the effectiveness and necessity of null-space projection in ensuring the mutual exclusivity of appearance and action features.
4) \textbf{Suppressing optimization shortcuts is crucial for retrieval performance.} Removing the gradient clipping operation (D\#7) leads to the most significant performance degradation (R@1 drops to 80.55\%). This indicates that the normal component during backpropagation will drive the model to seek harmful optimization shortcuts (Shortcut Learning), causing the model to forcefully explain action differences utilizing appearance differences. Blocking this shortcut is the key to preventing the action extraction rule from being tampered with.
5) \textbf{Uncertainty estimation promotes smooth optimization transitions.} In the GR module, replacing the adaptive weight with a fixed strategy that linearly increases with epochs (D\#10) causes R@1 to decrease to 83.92\%. This confirms that the dynamic confidence based on Shannon entropy can more accurately measure the current action semantic learning state of the model, thereby achieving a smoother gradient rectification transition.

\textbf{Environmental robustness under the Multi-Weather setting.}
Table~\ref{tab:ablation_lightair_full} shows the same trend under weather corruption: raw visual features (D\#8) and removing Stop-Gradient (D\#7) reduce mean R@1 to 64.32\% and 62.32\%, respectively, while Sigmoid activation (D\#3) reaches 66.93\%. These results support decoupled features, sparse probability modeling, and gradient rectification for robust alignment.

\textbf{Generalization on the OOD setting.}
We analyze the performance of the model on the open domain dataset UCC in Table~\ref{tab:ablation_lightair_ucc}:
1) \textbf{Orthogonal projection guarantees cross domain generalization.} Replacing null-space projection with simple feature subtraction (D\#5) or omitting the Top-K sparsification operation (D\#6) causes R@1 to drop to 60.00\% and 59.41\%, respectively. This indicates that linear operations easily cause dominant appearance information to recontaminate the action representation, while omitting the sparsification operation might introduce additional redundant action noise. Therefore, the action features reconstructed based on the sparse semantic basis, combined with the pure appearance features obtained via null-space projection, prompt the model to capture essential semantics with domain invariance, thereby significantly enhancing its cross domain generalization.
2) \textbf{Removing harmful optimization shortcuts is key to achieving cross domain generalization.} Removing the gradient rectification operation (D\#7) causes a significant decline in R@1 on the UCC test set to 53.48\%. This indicates that although the model might still overfit to specific appearance shortcuts in the original data, when facing distribution shifts in OOD scenes, it is imperative to extract invariant essential action semantics through strict gradient constraints to guarantee model generalization.
3) \textbf{Adaptive weights guarantee smooth optimization convergence.} Removing the Shannon entropy based adaptive transition strategy (D\#9) causes R@1 to decrease to 60.47\%. This indicates that when handling cross domain data, the action semantic manifold is highly unstable in the early training stages. Directly applying strict Riemannian rectification will cut off normal feature exploration paths and hinder convergence. Introducing a confidence based dynamic rectification mechanism to ensure smooth optimization transition is the key to ultimately achieving retrieval performance improvement.

\section{Algorithm of Training Procedure}
\label{sup:training_procedure}
To elucidate the end-to-end optimization of LightAIR, Algorithm~\ref{alg:lightair} summarizes its training process. First, utilizing the AIO module, we reconstruct the pure action feature $\mathbf{z}_{act}$ from coupled visual features by estimating latent action coefficients with the semantic codebook $\mathbf{D}$, and align it with the ground-truth action semantics $\mathbf{t}_y$. Next, the ONSP module constructs the action projection operator $\mathbf{\Pi}_{act}$ and projects visual features onto its null-space to strictly eliminate action information, thereby extracting the pure static appearance feature $\mathbf{z}_{app}$. An adaptive gating network then aggregates the dual-stream features into the final visual representation $\mathbf{z}_{final}$ to compute the multimodal retrieval loss. Finally, during the backpropagation phase, the GR module calculates the adaptive uncertainty weight $\omega$ based on the Shannon entropy of the action activation probability to dynamically rectify the original Euclidean gradient $\mathbf{g}_{Euc}$ into the Riemannian gradient $\mathbf{g}_{Riem}$. This strictly confines the gradient within the valid semantic subspace to cut off potential harmful optimization shortcuts. The model undergoes joint optimization by minimizing the composite objective $\mathcal{L}_{total}$ and updates parameters using the rectified gradient $\mathbf{g}_{final}$, simultaneously ensuring effective feature decoupling and semantic basis stability.

\begin{algorithm}[t]
\small
\caption{Algorithm of LightAIR's Training Procedure}
\label{alg:lightair}

\raggedright
\textbf{Input}: Image-text pair set $\mathcal{T} = \{(\mathbf{x}_\mathbb{T}, \mathbf{x}_\mathbb{I})_n\}_{n=1}^N$. \\
\textbf{Parameter}: Max epochs $N$, batch size $B$, temperatures $\tau$, trade-off hyperparameters $\gamma_1, \gamma_2$. \\
\textbf{Output}: Fine-tuned model.\\

\begin{algorithmic}[1]
\REQUIRE Image-text dataset $\mathcal{T}$, semantic codebook $\mathbf{D} \in \mathbb{R}^{L \times D}$, model parameters $\mathbf{\Theta}$

\FOR{epoch $I = 1$ to $N_{ep}$}
    \STATE Sample a batch $\mathcal{B} \subset \mathcal{T}$ of size $B$
    
    \STATE {// \textbf{Forward Pass: Feature Extraction \& Decoupling}}
    \FOR{each pair $(\mathbf{x}_\mathbb{T}, \mathbf{x}_\mathbb{I}) \in \mathcal{B}$}
        \STATE Extract global visual and query features:
        \\ $\quad \mathbf{v} \leftarrow \Phi_\mathbb{I}(\mathbf{x}_\mathbb{I}), \quad \mathbf{t}_q \leftarrow \Phi_\mathbb{T}(\mathbf{x}_\mathbb{T})$
        \STATE Extract ground-truth action semantic guidance:
        \\ $\quad \mathbf{t}_y \leftarrow \Phi_\mathbb{T}(\mathbf{x}_\mathbb{T})$
        
        \STATE {// \textit{1. Action Inversion Operator (AIO)}}
        \STATE Estimate latent action coefficient (Eq.~\ref{eq:topk}): $\boldsymbol{\alpha} \leftarrow \text{Softmax}(\text{Top-K}(\Phi_{act}(\mathbf{v})))$
        \STATE Reconstruct pure action feature via $\mathbf{D}$ (Eq.~\ref{eq:zact}): $\mathbf{z}_{act} \leftarrow \frac{\mathbf{D}^\top \boldsymbol{\alpha}}{\|\mathbf{D}^\top \boldsymbol{\alpha}\|_2}$
        \STATE Calculate action semantic alignment (Eq.~\ref{eq:cls}): $\mathcal{L}_{cls} \leftarrow 1 - \text{Cosine}(\mathbf{z}_{act}, \mathbf{t}_y)$

        \STATE {// \textit{2. Orthogonal Null-Space Projection (ONSP)}}
        \STATE Construct action projection operator(Eq.~\ref{eq:pi}): $\mathbf{\Pi}_{act} \leftarrow \frac{\mathbf{z}_{act} \mathbf{z}_{act}^\top}{\|\mathbf{z}_{act}\|^2 + \epsilon}$
        \STATE Extract pure appearance feature (Eq.~\ref{eq:nullspace}): $\mathbf{z}_{app} \leftarrow (\mathbf{I} - \mathbf{\Pi}_{act})\mathbf{v}$
        \STATE Compute adaptive gating weights (Eq.~\ref{eq:mlp}): $\mathbf{W} \leftarrow \text{MLP}([\mathbf{z}_{act}, \mathbf{z}_{app}])$
        \STATE Aggregate final visual feature (Eq.~\ref{eq:combine}): $\mathbf{z}_{final} \leftarrow \mathbf{W}_{act} \odot \mathbf{z}_{act} + \mathbf{W}_{app} \odot \mathbf{z}_{app}$
    \ENDFOR

    \STATE {// \textbf{Loss Computation}}
    \STATE Calculate image-text contrastive loss $\mathcal{L}_{itc}$ (Eq.~\ref{eq:itc}):
    \\ $\quad \mathcal{L}_{itc} \leftarrow - \frac{1}{2B} \sum_{i=1}^{B} \left( \log \frac{\exp(s_{i,i})}{\sum_{j=1}^B \exp(s_{i,j})} + \log \frac{\exp(s_{i,i})}{\sum_{j=1}^B \exp(s_{j,i})} \right)$
    \STATE Calculate image-text matching loss $\mathcal{L}_{itm}$ (Eq.~\ref{eq:itm}):
    \\ $\quad \mathcal{L}_{itm} \leftarrow - \frac{1}{B} \sum_{i=1}^{B} \Big( y_i \log p_m^i + (1 - y_i) \log (1 - p_m^i) \Big)$
    \STATE Compute global optimization objective (Eq.~\ref{optimization}): 
    \\ $\quad \mathcal{L}_{total} \leftarrow \mathcal{L}_{itc} + \gamma_1 \mathcal{L}_{itm} + \gamma_2 \mathcal{L}_{cls}$

    \STATE {// \textbf{Backward Pass: Gradient Rectification (GR)}}
    \FOR{each pair $(\mathbf{x}_\mathbb{T}, \mathbf{x}_\mathbb{I}) \in \mathcal{B}$}
        \STATE Obtain original Euclidean gradient (Eq.~\ref{eq:Euc}): $\mathbf{g}_{Euc} \leftarrow \nabla_{\mathbf{v}} \mathcal{L}_{total}$
        \STATE Project to Riemannian gradient (Eq.~\ref{eq:riemannian projection}): $\mathbf{g}_{Riem} \leftarrow (\mathbf{I} - \mathbf{\Pi}_{act})\mathbf{g}_{Euc}$
        \STATE Compute Shannon entropy (Eq.~\ref{eq:omega}): $\mathcal{H} \leftarrow - \sum_{l=1}^L \alpha_l \log \alpha_l$
        \STATE Calculate adaptive uncertainty weight (Eq.~\ref{eq:omega}): $\omega \leftarrow \exp(-\mathcal{H}/\tau)$
        \STATE Rectify final gradient (Eq.~\ref{eq:g_final}): $\mathbf{g}_{final} \leftarrow (1 - \omega)\mathbf{g}_{Euc} + \omega \mathbf{g}_{Riem}$
    \ENDFOR
    
    \STATE {// \textbf{Parameter Update}}
    \STATE Update model parameters $\mathbf{\Theta}$ via backpropagation using the rectified gradients $\{\mathbf{g}_{final}\}$
\ENDFOR
\end{algorithmic}
\end{algorithm}

\section{More Case Study}
\label{sup:more_case}

\begin{figure*}[ht]
    \centering
    \vspace{-10pt}
	\includegraphics[width=0.97\linewidth]{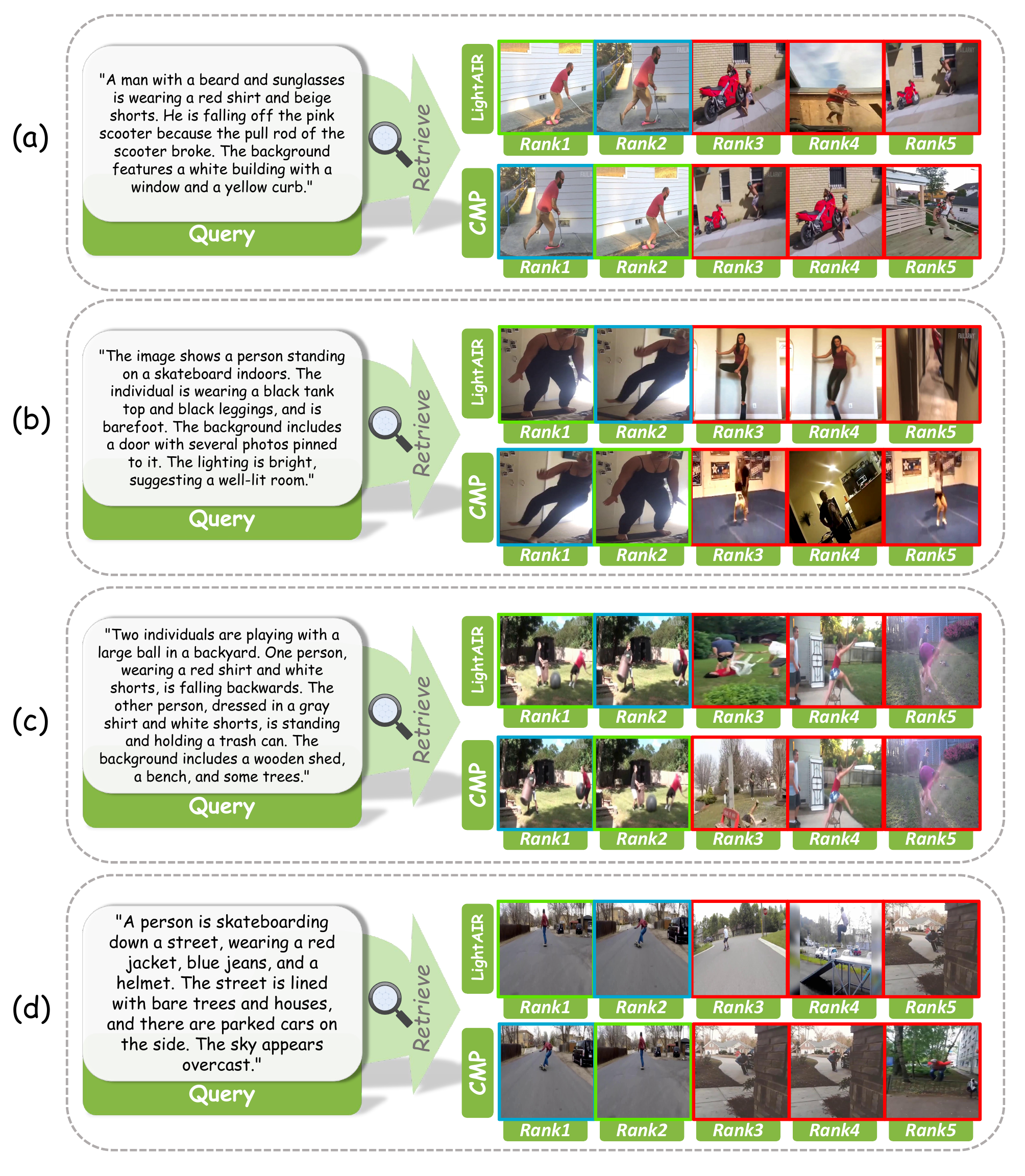}
      \vspace{-10pt}
	\caption{Qualitative comparison of successful retrieval cases between LightAIR and the baseline CMP. Matched images are marked by green boxes, mismatched images are marked in red, and blue boxes indicate hard negatives.}
	\label{fig:CaseImproved}
\end{figure*}

\begin{figure*}[ht]
    \centering
    \vspace{-10pt}
	\includegraphics[width=0.97\linewidth]{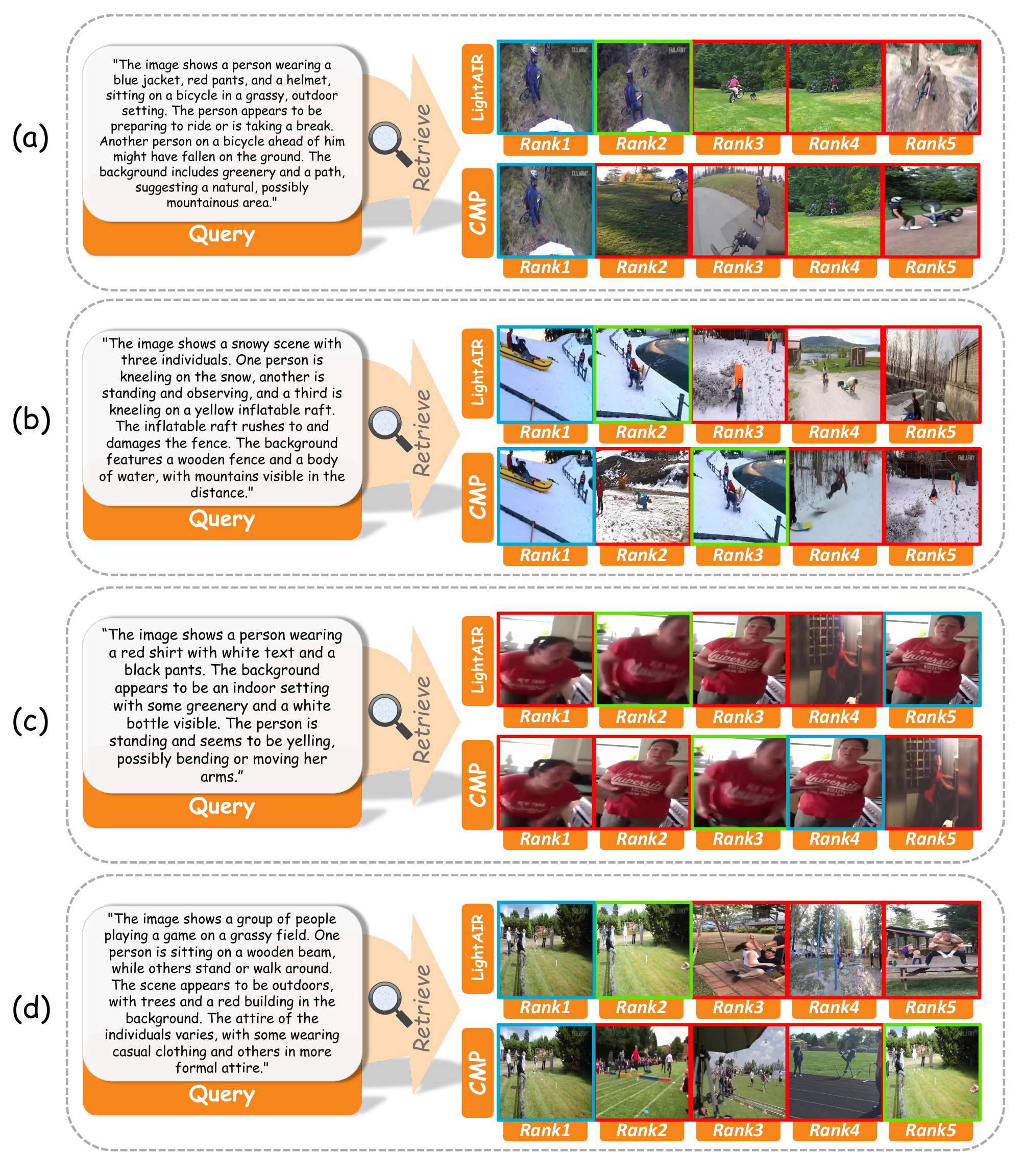}
      \vspace{-10pt}
	\caption{Qualitative analysis of failure cases for both LightAIR and CMP. Matched images are marked by green boxes, and mismatched images are marked in red, and blue boxes indicate hard negatives.}
	\label{fig:CaseFailure}
\end{figure*}

To further qualitatively validate the cross-modal retrieval performance of LightAIR, Figure~\ref{fig:CaseImproved} and Figure~\ref{fig:CaseFailure} respectively present comparisons of success and failure cases against the baseline model CMP across various complex scenarios.

As shown in Figure~\ref{fig:CaseImproved}, LightAIR demonstrates significant retrieval robustness when handling various highly deceptive hard negative samples:

Figure~\ref{fig:CaseImproved} (a) demonstrates the model's ability to overcome the explicit static appearance dominance issue. When retrieving ``falling off the pink scooter'', CMP's Top-1 result accurately matches explicit appearance features like ``red shirt'' and ``pink scooter'', but recalls a sample where the target is in a ``normal riding'' state. This intuitively reflects visual feature decoupling failure caused by pixel-level representation entanglement. In contrast, LightAIR successfully hits the target, confirming that by relying on the text semantic prior of AIO and the null-space projection of ONSP, the model achieves strict geometric decoupling of action and appearance, effectively eliminating interference from dominant appearance signals.

Figure~\ref{fig:CaseImproved} (b) reveals the model's robustness in extracting features under weak action signals. For ``standing on a skateboard indoors'', a static action lacking significant dynamic trajectories, CMP is easily misled by similar indoor backgrounds or static poses. However, LightAIR robustly hits the target. This proves that even when the action visual feature is weak, the model can still rely on the pure semantic subspace to suppress background noise and precisely extract and align the core action component.

Figure~\ref{fig:CaseImproved} (c) verifies the model's effectiveness in cutting off harmful optimization shortcuts. Facing a complex multi-person scenario involving someone ``falling backwards'' and another ``holding a trash can'', CMP generates severe false recalls. LightAIR, conversely, successfully overcomes visual interference and hits the correct target. This is attributed to the Riemannian gradient rectification mechanism of the GR module. During backpropagation, it severs the cross branch interference dominated by the normal gradient, blocking the harmful shortcut where the model attempts to distort the underlying mapping rule when facing hard negative sample penalization. By strictly constraining the update trajectory within the tangent space of the action semantic manifold, it forces the model to achieve precise identification of weak action differences.

Figure~\ref{fig:CaseImproved} (d) reflects the model's cross-modal alignment capability in complex open scenarios. The query describes ``skateboarding down a street'' accompanied by local appearance attributes like ``red jacket''. In a street scene filled with high-frequency background noise (e.g., trees, houses, vehicles), CMP recalls interference samples with mismatched actions. In contrast, LightAIR successfully locates the target matching both action and appearance among numerous distractors. This further confirms that the model can construct clearer and more robust cross-modal semantic decision boundaries under open-world noise.

Figure~\ref{fig:CaseFailure} explores the limitations and failure scenarios of the current model when handling extremely hard samples:

Figure~\ref{fig:CaseFailure} (a) reveals the feature attenuation issue caused by distant small targets. For queries describing a distant target ``fallen on the ground'', the action signal is severely attenuated due to the extremely small pixel proportion of the person in the frame. At this extreme scale, even with the introduction of text semantic prior, it remains difficult for model to extract effective action anchors from underlying visual features, causing cross-modal alignment failure.

Figure~\ref{fig:CaseFailure} (b) reflects the modeling bottleneck of complex physical interactions. Retrieving ``inflatable raft... damages the fence'' involves composite interactions among people, objects, and the environment, accompanied by severe occlusion and non-standard postures. Because such samples severely deviate from the regular action semantic manifold, the current semantic codebook reconstruction mechanism based on the AIO module remains insufficient to fully deconstruct these complex physical interaction semantics.

Figure~\ref{fig:CaseFailure} (c) demonstrates the deficiency in fine-grained local action perception. For queries like ``yelling'' that lack large-scale spatial motion trajectories and rely solely on facial or local micro-actions, weak local action features are easily overshadowed by explicit large-area appearance attributes (e.g., ``red shirt''). This indicates that the current forward decoupling mechanism still faces the risk of interference from dominant appearance information when handling extremely fine-grained local actions.

Figure~\ref{fig:CaseFailure} (d) explores static posture ambiguity within dense crowds. Retrieving ``sitting on a wooden beam'' against a background containing a dense crowd to ``stand or walk around'' is highly challenging for the fine-grained distinction of highly similar static postures lacking dynamic clues. Under severe visual confusion, it is difficult for the global semantic prior to establish reliable feature mapping at the pixel scale.

